# AscDAMs: Advanced SLAM-based channel detection and mapping system

Tengfei Wang[1†], Fucheng Lu[1†], Jintao Qin[1], Taosheng Huang[1], Hui Kong[2]*, Ping Shen[3]*

[1] State Key Laboratory of Internet of Things for Smart City and Department of Civil and Environmental Engineering, University of Macau, Macao SAR, People's Republic of China

[2] The State Key Laboratory of Internet of Things for Smart City and Department of Electromechanical Engineering, University of Macau, Macao SAR, People's Republic of China

[3] State Key Laboratory of Internet of Things for Smart City and Department of Ocean Science and Technology, University of Macau, Macao SAR, People's Republic of China

[†]These authors contributed equally to this work and should be considered co-first authors

*Corresponding authors. Ping Shen: pingshen@um.edu.mo; Hui Kong: huikong@um.edu.mo

**Abstract**: Obtaining high-resolution, accurate channel topography and deposit conditions is the prior challenge for the study of channelized debris flow. Currently, wide-used mapping technologies including satellite imaging and drone photogrammetry struggle to precisely observe channel interior conditions of mountainous long-deep gullies, particularly those in the Wenchuan Earthquake region. SLAM is an emerging tech for 3D mapping; however, extremely rugged environment in long-deep gullies poses two major challenges even for the state-of-art SLAM: (1) Atypical features; (2) Violent swaying and oscillation of sensors. These issues result in large deviation and lots of noise for SLAM results. To improve SLAM mapping in such environments, we propose an advanced SLAM-based channel detection and mapping system, namely AscDAMs. It features three main enhancements to post-process SLAM results: (1) The digital orthophoto map aided deviation correction algorithm greatly eliminates the systematic error; (2) The point cloud smoothing algorithm substantially diminishes noises; (3) The cross section extraction algorithm enables the quantitative assessment of channel deposits and their changes. Two field experiments were conducted in Chutou Gully, Wenchuan County in China in February and November 2023, representing observations before and after the rainy season. We demonstrate the capability of AscDAMs to greatly improve SLAM results, promoting SLAM for mapping the specially challenging environment. The proposed method compensates for the insufficiencies of existing technologies in detecting debris flow channel interiors including detailed channel morphology, erosion patterns, deposit distinction, volume estimation and change detection. It serves to enhance the study of full-scale debris flow mechanisms, long-term post-seismic evolution, and hazard assessment.

# 1 Introduction

The 2008 Wenchuan Earthquake produced a huge amount of loose solid material, spawning repeated post-seismic channelized debris flows in numerous long-deep gullies (Tang et al., 2009; Guo et al., 2016; Fan et al., 2019), which have posed a continuing threat to human lives and properties (Xu et al., 2012; Hu & Huang, 2017; Zhang & Zhang, 2017; Fan et al., 2018; Fan et al., 2019; Shen et al., 2020; X. Z. Zhang et al., 2022; Zhang et al., 2023). During the 15 years after the earthquake, the hillslope loose material have been transported gradually into channel deposit (Zhang & Zhang, 2017), resulting in significantly change in the occurrence frequency and initiation mechanisms of debris flow (Berger et al., 2011b; Chen et al., 2024). The channel morphology continually changes with the process of sediment transportation. The morphology of debris flow channel indicates the channel topography elevation, loose material distribution, debris flow impact area, entrainment area and entrainment depth, deposition area and volume, etc. (Remaître et al., 2005). An accurate observation of debris flow channel morphology is vital to the deep understanding of debris flow initiation mechanism and risk assessment. Besides, the accuracy of many debris flow numerical simulation frameworks, e.g., r.avaflow (Mergili et al., 2017) and EDDA (Chen & Zhang, 2015; Shen et al., 2018), and debris flow assessment factors (Liang et al., 2012; Meyer et al., 2014; Li et al., 2021) are highly relied on the accuracy of topography data.

Conventionally, morphology data of debris flow channels are mainly obtained through satellite images and field investigation. Satellites can easily produce digital elevation model (DEM) and digital orthophoto map (DOM) (Zhang et al., 2014; Mueting et al., 2021; Luo et al., 2022) of a wide range of area (over tens of square kilometers) with low average unit cost. However, satellite-derived DEM and DOM often exhibit limited resolution and accuracy in rugged terrains especially in mountainous area and deep valleys (Sun et al., 2015; Zhou et al., 2017; Liu et al., 2021). Field investigation by visual observation and manual measurement by ruler or laser range finder could be less efficient and imprecise. Recently, equipment like radar (Schurch et al., 2011; Caduff et al., 2015; Morino et al., 2019; Bonneau et al., 2022) and unmanned aerial vehicle (UAV) (Simoni et al., 2020; Walter et al., 2022; X. Z. Zhang et al., 2022; Zhang et al., 2023) have been widely applied in field investigation to get more detailed topographic maps. However, these methods are still restricted by the application environment. For example, UAV mapping requires open airspace, enough signal strength of GNSS as well as skilled operators (Cucchiaro et al., 2019; Imaizumi et al., 2019; Huang et al., 2022). It is extremely difficult to operate in the space-narrowed, signal-blocked, and GNSS-denied deep valley environment. The radar-based approaches require an appropriate arrangement for installation locations and/or scanned areas that is also very difficult in this rugged and rocky region (Blasone et al., 2014; Caduff et al., 2015; Morino et al., 2019; Tang et al., 2022). Hence, current technologies could not provide sufficient, accurate and consistent information in deep valleys of alpine areas, where channelized debris flows initiated and developed, especially Wenchuan Earthquake region. These constraints impede a thorough understanding of the debris flow mechanisms. Therefore, developing a new method for accurately detecting debris flow channels is an urgent, common key issue in channelized debris flow research and hazard mitigation.

Simultaneous localization and mapping (SLAM) (Bailey & Durrant-Whyte, 2006; Durrant-Whyte & Bailey, 2006; Cadena et al., 2016; Barros et al., 2022) technology is a mobile measurement method that continuously records data on the move. It has a wide usage in the field of robotic navigation, autonomous driving, and topographic mapping. One typical kind of SLAM technology is LIDAR odometry and mapping (LOAM) (Zhang & Singh, 2014) based on light radar (LIDAR), which matches point clouds with their features. LOAM distinguishes features based on the curvature of points scanned by LIDAR. By this method, the computational complexity can be reduced. SLAM technology has been applied across different platforms for numerous scenarios of topographic mapping including forestry (Kukko et al., 2017; Pierzchala et al., 2018; Li et al., 2023), underground tunnel (Ullman et al., 2023), urban morphology (Tanduo et al., 2022), hillslope gullies (Kinsey-Henderson et al., 2021), and densely vegetated hillsides (Marotta et al., 2021). SLAM has excellent potential in supplementing the data of debris flow channel conditions where the satellite images are of low quality.

Nonetheless, it is still challenging to apply state-of-art SLAM techniques in debris flow gullies in alpine area, e.g., Wenchuan Earthquake region, from in-situ preliminary tests. **Long-deep debris flow gully with extremely rugged environment have two major challenges for successful application of SLAM** (see visual details in Section 2): **(1) Atypical features.** The channelized debris flow gullies are typically long and deep. Channel sidewalls are sparsely vegetated while the channel bed is bestrewed with loose materials. The features of this specific environment are atypical. It is difficult to acquire sufficient and effective channel morphology data. **(2) Violent swaying and oscillation of sensors.** Due to the presence of large rocks and flowing streams within the channel, activities like climbing, jumping, and rotating are unavoidable during the data collection process. Hence, the sensors experience inevitable severe swaying and rocking. These issues result in: (1) The SLAM algorithm was apt to produce large mapping deviation because the algorithm is unable to extract enough effective features for matching and computing in this environment with atypical features. Besides, SLAM algorithms relying on scanning and matching of point cloud frames will lead to systematic tiny errors that accumulate into considerable deviation with the increase of channel length and the expand of data set. Although many efforts have been made to mitigate the influence of the deviation, for example, introducing visual-inertial odometry, looper closure and GNSS information for pose correction (Shan et al., 2020; Lin & Zhang, 2022a), the mapping deviation is still prominent for such long-deep gullies. (2) Due to the significant oscillation of sensors on the backpack-type collection system during mountaineering process, the mapping result of SLAM algorithm contains lots of noise.

The above limitations have hindered the application of current SLAM technology in investigating debris flow channels. To address these problems, we propose an advanced SLAM-based channel detection and mapping system (AscDAMs), which contains three major novel contributions to post-process SLAM results.

(1) Deviation correction algorithm. The newly proposed algorithm uses the DOM and barometric elevation data as a benchmark to minimize the accumulation drift, offering a viable solution to the morphology detection challenge in long and deep channels.

(2) Point cloud smoothing algorithm. This algorithm is designed to mitigate the noise caused by drastic motion state change. It has the capability to enhance the quality of SLAM results, even in scenarios where sensors undergo violent oscillations during data detection in the harsh channel environment.

(3) Cross-section extraction algorithm. This is beneficial for extracting typical channel cross-section profiles. This functionality enables the quantitative assessment of the channel interior including the channel deposits distinction, volume estimation, change monitoring and erosion pattern observation, etc.

In the following content, the algorithms and application process of AscDAMs will be introduced in detail. The outputs of AscDAMs are demonstrated to have huge potential to greatly propel the comprehensive exploration of the underlying mechanisms of debris flow and the long-term evolution of debris flow activities.

## 2  Application environment

As one of the typical post-seismic debris flow gullies in Wenchuan County, the Chutou Gully is featured by steep hillslopes and deep channel (see Figure 1a-b). The disastrous earthquake on 12 May 2008 provided a huge amount of loose material inside the Gully that can be easily mobilized into debris flows (Tang et al., 2009; Guo et al., 2016; Yang, Tang, Cai, et al., 2023). Three large-scale debris flows occurred in the Chutou Gully catchment during the rainy season in 2013, 2019, and 2020 (X. Z. Zhang et al., 2022; Zhang et al., 2023). The repeated debris flow hazards not only pose a continuing threat to local people's lives and properties but also change the channel's and Minjiang River's morphology completely (see Figure 1c). After the earthquake, most of the loose material on the hillslopes have been moved down into the channel (Xiong et al., 2022; Y. Y. Zhang et al., 2022; Zhang et al., 2023). A recent conceptual model of post-seismic debris flow shows that the debris flow activity in the Wenchuan earthquake region is still at the unstable stage, showing a relatively high future risk of debris flow (Yang, Tang, Tang, et al., 2023). This makes it imperative to know the channel interior conditions for mechanisms study, debris flow prediction, and risk assessment.

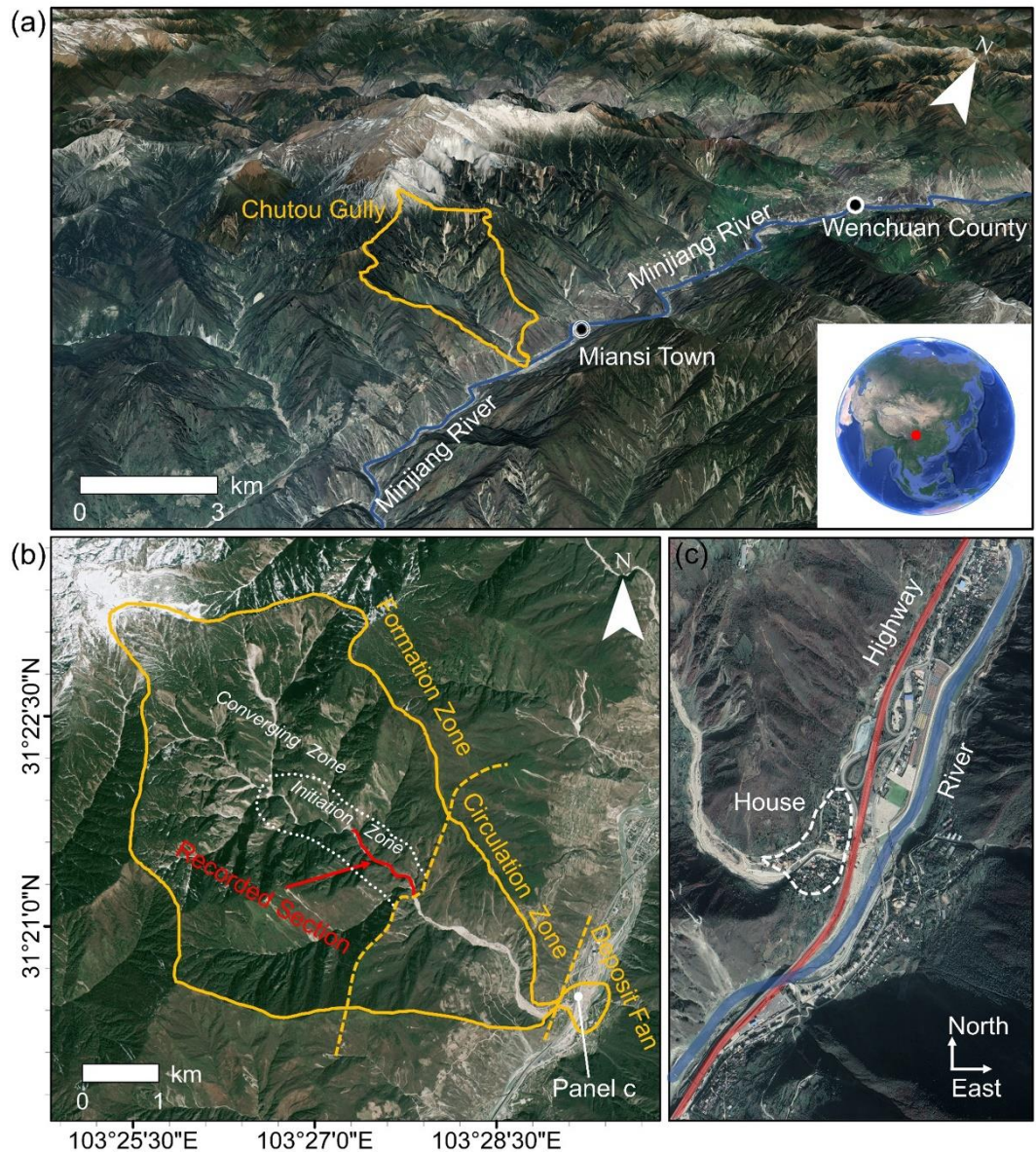

*Figure 1. (a) 3D map of the mountainous area of Wenchuan on 14 November 2021 from Google Earth. (b) Overview of Chutou Gully on 25 March 2020 from ZY03 satellite image. (c) Accumulation zone on 14 November 2021 from Google Earth.*

The gully can be divided into formation zone, circulation zone and deposit fan (Figure 1b). The formation zone is about 16 km² with complicated topographical conditions. We further partition the formation zone into a converging zone and an initiation zone. The converging zone is mainly exposed hillslopes for runoff convergence which can be detected directly by satellites. Recent debris flows, occurring in 2019 and 2020, originated in the initiation zone (X. Z. Zhang et al., 2022; Zhang et al., 2023). Loose materials distributed in this zone play a pivotal role in initiating and amplifying debris flows. In downstream, the channel in circulation zone is broad, and the terrain in the deposit fan is gentle, making detection easy by satellites or UAVs. However, different from these zones, the initiation zone is characterized by high-rising, inward-sloping, overhanging channel side walls as indicates by Figure 2. The overhanging cliffs obstruct the overhead view, and GNSS signal blockage by high mountains makes it channel to obtain accurate and detailed channel interior morphology by satellites and UAVs. Hence, we selected the initiation zone as the study area for the current research to showcase the novelty of the proposed approach.

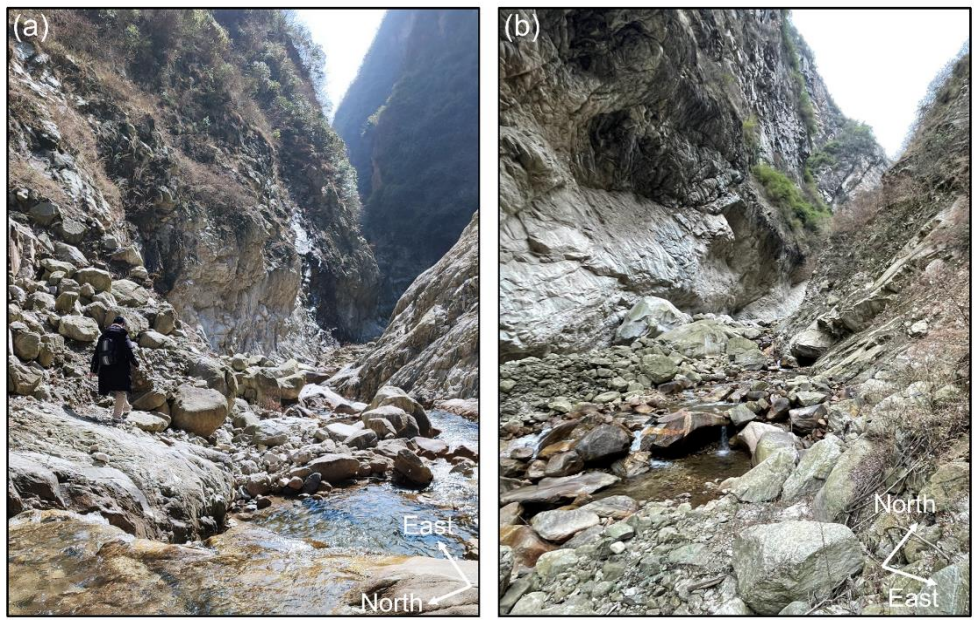

*Figure 2. Environment inside the Chutou Gully on 10 February 2023.*

SLAM algorithms were mainly applied to relatively stable environments, e.g., urban roads, river courses, park trails, etc. They perform satisfactorily in these smooth environments (Ye et al., 2019; Shan et al., 2020). However, as depicted in Figure 2, the channel bed is covered with loose deposits, rendering the channel path an exceeding rugged terrain. The rocky cliffs on both sides of the channel are uneven. A natural stream exists inside the channel, but the water surface is also unsteady due to the channel gradient and bumpy bed. These atypical channel features make it particularly harsh for the implementation of SLAM algorithms. The long and deep channel morphology, together with the complex channel environment, extend the working duration of SLAM platforms, e.g., handheld, backpack, helmet, etc., consequently enlarging the size of the dataset. This, in turn, will introduce additional errors to the results of SLAM. During the data acquisition, the huge stones and stream have emerged as obstacles, introducing difficulties to SLAM detaction. The abrupt changes in sensor pose caused by climbing and jumping, coupled with sudden swerving, can result in inaccurate pose estimations, generating a significant amount of noise consequently. AscDAMs is thus proposed to conquer these technical challenges for SLAM.

## 3 Methodology

### 3.1 Equipment and Data

The data acquisition system (Figure 3) of AscDAMs is backpack-type and consists of a LIDAR, an inertial measuring unit (IMU), a small-form-factor computer, a camera, etc. It is cost-effective and lightweight. The total weight of the device together with the battery is lower than 10 kg, with the main cost not exceeding 10,000 USD. Detailed information about its core configuration is shown in Table 1. We choose the LIDAR with 16 scan lines, the IMU with a frequency of 400 Hz, and the camera with a 1280×720 resolution. When assembling, it is important to note that the LIDAR and IMU must be fixed on the bracket to avoid unexpected errors. The view of LIDAR and camera should be unobstructed. The LIDAR and IMU offer data for SLAM to calculate the map, and the camera is designed to provide color information for channel characteristic extraction, e.g., channel deposit, vegetation, etc. We utilized the system to record data of the selected zone on 10th February 2023, with the recorded section more than 1-km long (shown in Figure 1b). A repeated channel scanning was conducted on 8th November 2023. It is noted that the second investigation is shorter than the first one because we were blocked by the rising water. The data characteristics are shown in Table 2.

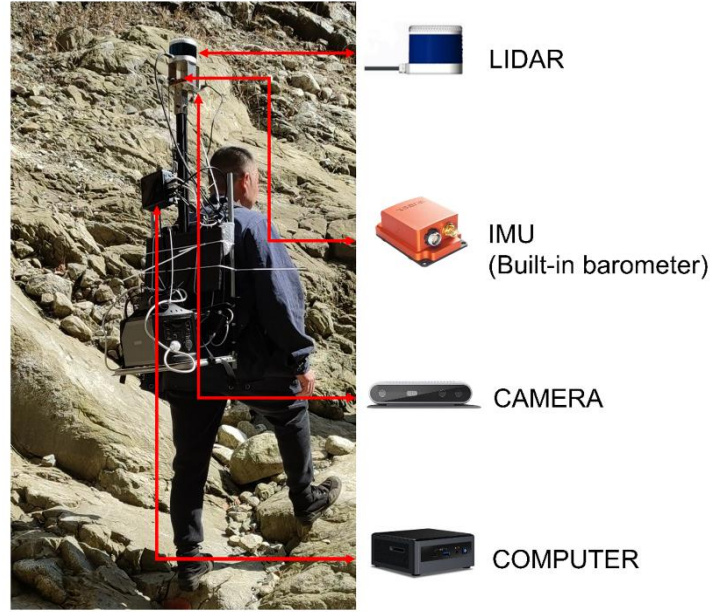

*Figure 3. The photo of the data acquisition system of AscDAMs.*

*Table 1. The list of core devices and their parameters.*

| Device | Model | Performance |
|---|---|---|
| LIDAR | Leishen C16 | Channel Number: 16<br>Max Data Rate: 320,000 Points/Second<br>Ranging Accuracy: ± 3 cm<br>Rotation Frequency: 5 Hz<br>Field of View: Horizontal: 360°; Vertical: -15°~15°<br>Angular Resolution: Horizontal: 2°; Vertical: 0.09° |
| IMU | Xsens Mti-G-710 | Navigation Accuracy: Roll/Pitch: 0.2° RMS; Yaw: 0.8° RMS<br>Velocity Accuracy: 0.05 m/s<br>Frequency: 400 Hz |
| CAMERA | Intel RealSense D415 | Image Resolution: 1280 × 720<br>Frames per Second: 30<br>Field of View: Horizontal: 69°; Vertical: 42° |
| COMPUTER | Intel NUC10FNH | Processor: Intel i7-10710U with CPU 1.10 Ghz × 12<br>Memory: 62.5 GiB<br>Disk: 2.0 TB<br>Operating System: Linux Ubuntu 20.04 |

*Table 2. The characteristics of the recorded data in Chutou Gully.*

| Item | Characteristic | |
|---|---|---|
| Date | 10th Feb. 2023 | 8th Nov. 2023 |
| Duration | 3,280 seconds | 1,300 seconds |
| Size | 49.4 GB | 19.4 GB |
| LIDAR point cloud | 16,177 messages | 6,447 messages |
| IMU data | 1,312,046 messages | 520,381 messages |
| Camera image | 97,629 messages | 38,721 messages |
| Pressure | 164,003 messages | 65,048 messages |

**3.2 Algorithms**

The entire workflow (Figure 4) of AscDAMs consists of the application of SLAM and several newly proposed algorithms including deviation correction, point cloud smoothing, cross-section extraction, and map coloring. First, SLAM registers and fuses LIDAR point cloud frames with motion state (i.e., the acceleration and pose of LIDAR) recorded by IMU to generate an initial global map and an initial trajectory of data acquisition. Then, we introduced a new algorithm for deviation correction of the initial global map, called DOM-and-barometer-aided deviation correction (DBADC). The coordinate of each point in the global map and trajectory will be corrected by referring to the horizontal coordinate of DOM and elevation of barometer. Next, the modified global map will be smoothed utilizing a new algorithm, called weighted-elastic voxel grid (WEVG). Finally, cross section extraction is processed, including cutting and projecting the smoothed global map, and smoothing and densifying cross sections. The final global map containing coordinate information can be converted to DEM. In addition, a map coloring algorithm is developed to supplement more information to the global map. It fuses camera RGB data with LIDAR data and expands SLAM to obtain colored maps. The final-colored maps are obtained after color optimization.

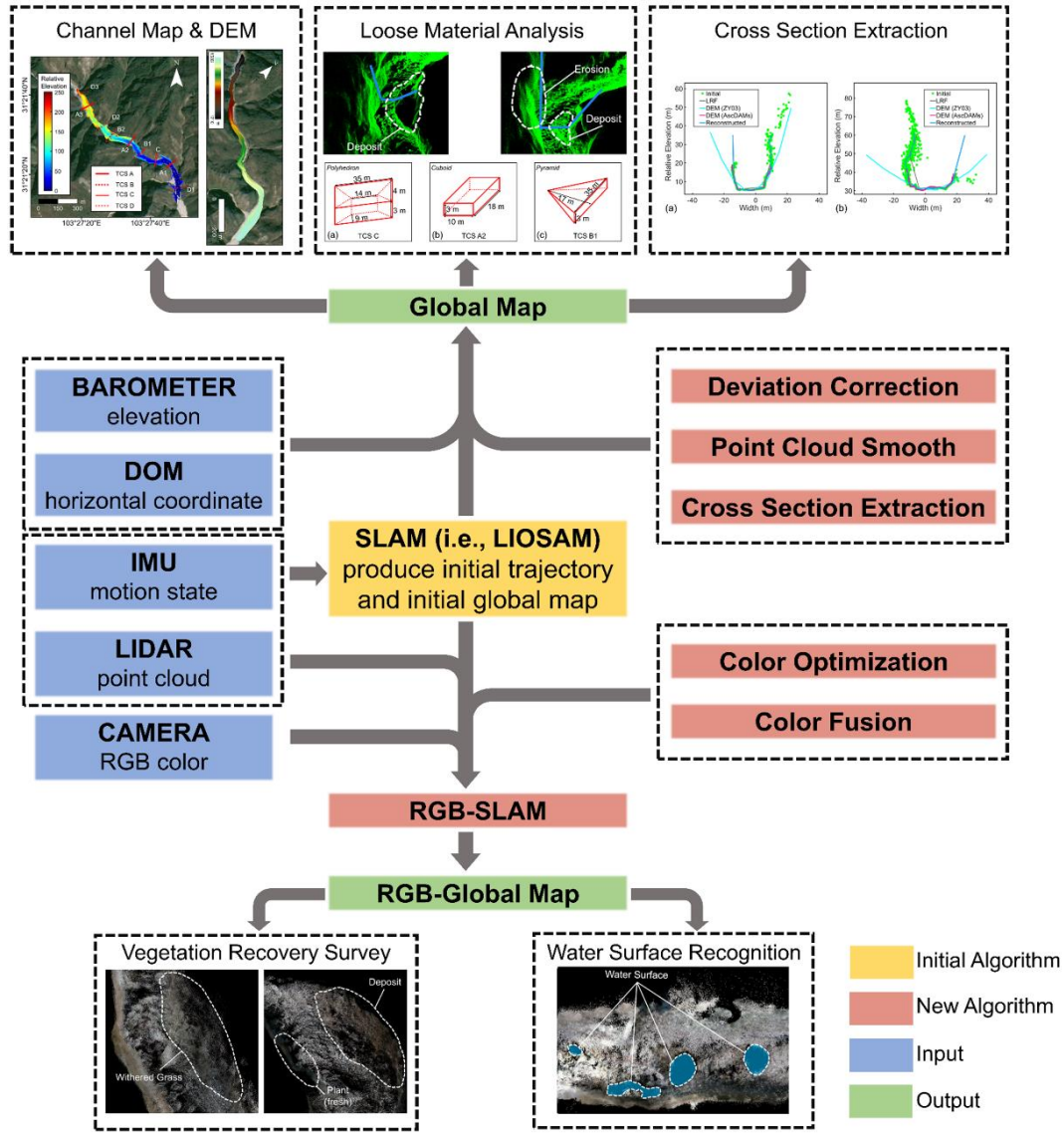

*Figure 4. Algorithm flow scheme of AscDAMs.*

3.2.1 Application of LIOSAM and Map Coloring

As one of the advanced LOAM algorithms, LIOSAM (Shan et al., 2020) merges with multiple sensors and enjoys the benefits of being accurate, open-source and easy to replicate which was widely used in 3D mapping. Therefore, this study is based on LIOSAM. Nonetheless, the choice of SLAM algorithm is not exclusive. AscDAMs is a universal channel morphology data acquisition and processing system, which is open to other LOAM algorithms. The success of applying LIOSAM to calculate the initial global map and trajectory map is a vital premise, as the subsequent optimization focusses on the two maps. For rugged debris flow channels, after various preliminary tests, it was found that the horizontal resolution of the input LIDAR point cloud images and the loop closure parameters have the greatest impact on whether we can obtain the maps. We achieved the application of LIOSAM in rugged environments by down sampling LIDAR point cloud and employing loop closure as iterative closest point (ICP) to rematch point clouds.

The initial LIOSAM map contains no color information while other SLAM algorithms have complex operations and strict requirements of equipment (Lin & Zhang, 2022b; Zheng et al., 2022). Hence, we propose a procedure to get colored maps of the channel. The procedure consists of three steps, namely color fusion, RGB-SLAM, and color optimization.

The method of color fusion refers to an existing algorithm of fusing camera and LIDAR on https://github.com/KAI-yq/camera_lidar_fusion (GitHub, 2021). We modify its way of coloring from point-by-point to frame-by-frame. This algorithm can fuse each LIDAR point cloud frame with simultaneous camera picture to get a single-frame colored point cloud, by calibrating and synchronizing the camera and LIDAR. RGB-SLAM is newly proposed to calculate the colored map in the present study. It is developed by modifying the data acquisition interface of LIOSAM. The implementation of RGB-SLAM is similar to that of LIOSAM.

However, there are many uncolored points in the initial map $C = \{c_1, c_2, c_3, \cdots, c_i, \cdots, c_I\}$ calculated by RGB-SLAM because of the limitation of the camera view. Hence, we have developed an algorithm to optimize the color of the map. First, the uncolored points in $C$ are filtered to produce a new colored map $C' = \{c'_1, c'_2, c'_3, \cdots, c'_j, \cdots, c'_J\}$ as coloring reference. Then, the color data around the uncolored point is searched out employing KNN. The searched color is given to the untouched point. Finally, points that still do not get color after re-coloring are filtered out, and the final colorful map $C'' = \{c''_1, c''_2, c''_3, \cdots, c''_k, \cdots, c''_K\}$ is completed.

3.2.2 Deviation Correction

The initial maps produced by LIOSAM have large deviations due to the accumulation of atypical features matching errors and systematical errors. These errors can cause the SLAM-generated map to drift from its actual position. Hence, it is necessary to correct the initial mapping to mitigate the drift. We introduced a new algorithm to correct the mapping deviation referring to the horizontal coordinate of DOM and elevation of barometer, called DOM-and-barometer-aided deviation correction (DBADC).

Theoretically, the initial maps, which contain stereo spatial data, need to be modified in XYZ dimensions. However, for rotating LIDARs, the errors in the horizontal (i.e., XY) directions are the same and can be adjusted together. Hence, the proposed deviation correction algorithm introduces two scaling factors named horizontal scaling factor $f_h$ and elevation scaling factor $f_e$ to rectify the drift in the initial maps. Since GNSS data is not available in such gully environment, we estimate $f_h$ with satellite DOM and calculate $f_e$ by converting the air pressure values recorded by barometer embedded in IMU to altitude. The scaling factors indicate the degree of modification between the initial point cloud map and the real channel morphology data. The closer the factors are to 1, the smaller the adjustment will be.

We develop an auto-estimation algorithm to determine $f_h$. Firstly, an adaptive segmentation method (Bradley & Roth, 2007) is employed to automatically extract debris flow channel area $\Omega$ from the DOM. Then the channel can be fitted as a line $T_{DOM}$ by removing noise and curve fitting. After comparing every fragment of $T_{DOM}$ with the initial trajectory $T = \{t_1, t_2, t_3, \cdots, t_i, \cdots, t_I\}$ and calculating their correlation, we can find out the most relevant fragment $T'_{DOM}$ to $T$ (Equation 1). $T'_{DOM}$ can be regard as the correction benchmark of $T$. Finally, we utilize ICP method to determine $f_h$ (Equation 2). After scaling the trajectory $T$ with ratio variable $s$, who makes the difference between the scaled trajectory and $T'_{DOM}$ the smallest equal to $f_h$. The elevation scaling factor $f_e$ can be determined by referring to the elevation of two real locations where the data acquisition system starts and stops recording (named "referred start point" and "referred end point" respectively), shown as Equation 3. $\Delta L_e$ represents the elevation difference between these two points in the initial trajectory of the point cloud map, while $L_e$ is their real elevation variation converted by barometer data.

$$T'_{DOM} = \underset{T_{DOMfragment} \in T_{DOM}}{argmax} [(T^* \cdot T_{DOMfragment})/(\|T\| \cdot \|T_{DOMfragment}\|)] \quad (1)$$

$$f_h = \underset{s}{argmin}[sT - T'_{DOM}] \quad (2)$$

$$f_e = L_e/\Delta L_e \quad (3)$$

In deviation correction of the initial global map, we group the points within the same spatial interval on the initial global map into the same correction unit. The coordinate deviations of all points within the same unit are considered to be the same. The shorter the interval length, the more accurate the correction result. There are two hypotheses for the units:

(1) The construction and size of each unit are accurate. All the points in the same unit should be regarded as a whole and translated together when rectifying.

(2) The overall errors of the initial global map are evenly distributed across the units.

The rationality of the two hypotheses is that we have maximized the performance of LIOSAM by down sampling and employing loop closure as enhanced ICP registration (see Section 3.2.1). The interval length of the unit is short enough so that its error could be neglected. However, the cumulative deviation by numerous units should be treated seriously.

We first smooth the trajectory $T$ by employing Fourier transform (and/or other smooth methods) $\hat{f}$ to remove jitter. Then it is densified by linear interpolation $\hat{L}$ to obtain the smoothed-densified trajectory $T' = \{t'_1, t'_2, t'_3, \cdots, t'_j, \cdots, t'_J\}$ (Equation 4).

$$T' = \hat{L}(\hat{f}[T]) \quad (4)$$

After projecting $T'$ onto the horizontal (XY) plane, we get new trajectory point cloud $T'' = \{t''_1, t''_2, t''_3, \cdots, t''_j, \cdots, t''_J\}$ (Equation 5), which can be considered as a division index trajectory.

$$t''_j = t'_j \cdot \begin{bmatrix} 1 & 0 & 0 \\ 0 & 1 & 0 \\ 0 & 0 & 0 \end{bmatrix} \quad (5)$$

The division of the initial global map is the key step of deviation correction. The adjacent points should be grouped into the same or adjacent unit. Next, we employ K Nearest Neighbor method (KNN) (Ram & Sinha, 2019) to find the nearest trajectory point $t''_j$ on $T''$ for each point of the global map $G = \{g_1, g_2, g_3, \cdots, g_k, \cdots, g_K\}$. Points of $G$ who share the same nearest point are divided into the same unit $U_j = \{u_{j1}, u_{j2}, u_{j3}, \cdots, u_{jm}, \cdots, u_{jM}\}$ and mark with the same index $j$. It is noted that the number $M$ of points of each unit is unequal. Finally, we reorder all points of the global map according to the index $j$ to ensure that the points within the same unit are sequentially adjacent.

The point $t'_j$ of the trajectory $T'$ on which recombining different points of the global map to the same unit $U_j$ can be used to determine the unit's translation vector $\boldsymbol{A_j}$. The modified trajectories $\check{T} = \{\check{t}_1, \check{t}_2, \check{t}_3, \cdots, \check{t}_i, \cdots, \check{t}_I\}$ (Equation 6), $\check{T}' = \{\check{t}'_1, \check{t}'_2, \check{t}'_3, \cdots, \check{t}'_j, \cdots, \check{t}'_J\}$ (Equation 7), and $\check{T}'' = \{\check{t}''_1, \check{t}''_2, \check{t}''_3, \cdots, \check{t}''_j, \cdots, \check{t}''_J\}$ (Equation 8) can be calculated directly by multiplying coordinates of all trajectory points by scaling factors.

$$\check{t}_i = t_i \cdot \begin{bmatrix} f_h & 0 & 0 \\ 0 & f_h & 0 \\ 0 & 0 & f_e \end{bmatrix} \tag{6}$$

$$\check{t}'_j = t'_j \cdot \begin{bmatrix} f_h & 0 & 0 \\ 0 & f_h & 0 \\ 0 & 0 & f_e \end{bmatrix} \tag{7}$$

$$\check{t}''_j = t''_j \cdot \begin{bmatrix} f_h & 0 & 0 \\ 0 & f_h & 0 \\ 0 & 0 & f_e \end{bmatrix} \tag{8}$$

The translation vector $\boldsymbol{A_j}$ is equal to the coordinate difference of the trajectory point before and after modification (Equation 9). After moving all the points in the unit with the same translation vector $\boldsymbol{A_j}$, we attain the corrected unit $\widetilde{U}_j = \{\widetilde{u}_{j1}, \widetilde{u}_{j2}, \widetilde{u}_{j3}, \cdots, \widetilde{u}_{jm}, \cdots, \widetilde{u}_{jM}\}$ (Equation 10) and the adjusted global map $\widetilde{G}$ which is the sum of all $\widetilde{U}_i$ (Equation 11).

$$\boldsymbol{A_j} = \check{t}'_j - t'_j \tag{9}$$

$$\widetilde{u}_{jm} = u_{jm} - \boldsymbol{A_j} \tag{10}$$

$$\breve{G} = \{\breve{U}_1, \breve{U}_2, \breve{U}_3, \cdots, \breve{U}_j, \cdots, \breve{U}_J\} \tag{11}$$

3.2.3 Point Cloud Smoothing

The modified global point cloud $\breve{G} = \{\breve{g}_1, \breve{g}_2, \breve{g}_3, \cdots, \breve{g}_k, \cdots, \breve{g}_K\}$ is blurry and has numerous noises because of inevitable oscillations during data collection in the uneven channel. This poses a major obstacle to the observation of internal channel topography, and the identification and analysis of channel deposits. Based on Voxel Grid (VG) filtering method (Rusu & Cousins, 2011), we propose a new smoothing algorithm, weighted-elastic voxel grid (WEVG), which expands VG by adding weight factor and making the voxel size flexible.

The VG filter creates a 3D voxel raster from the input point cloud, then calculates the centroid of each voxel utilizing all points within it. All points within the same voxel are only represented by the centroid. VG filtering can preserve the point cloud's geometric structure (Liu & Zhong, 2014). It is suitable for point cloud maps of debris flow channel. However, the disadvantages of VG filtering cannot be neglected:

(1) The calculation of centroid is inaccurate without considering point density. In the precise computation of the centroid, density is an independent variable and will affect calculation in an inhomogeneous case.

(2) The size of each voxel is the same and fixed so that it is stiff to balance the elimination of outliers and the maintenance of details. The ideal voxel raster should be flexible to partition larger voxel where points are sparse and smaller voxel where points are dense.

The point distribution in the global map is inhomogeneous in this study. The map can be more satisfactory after inserting point density as a weight to the centroid calculation and transforming voxel size adaptive to the distribution of point clouds. We express $\rho_k$ (Equation 12), the density in point $\breve{g}_k$, as the reciprocal of volume $V_k$ taken up by it, where the volume is defined by applying KNN to search $\breve{g}_k$'s several nearest neighbors and calculating their average distance $\bar{r}_k$.

$$\rho_k = 1/V_k = 1/\{(4/3)\pi \bar{r}_k^3\} = 3/(4\pi \bar{r}_k^3) \tag{12}$$

The voxel is divided based on the principle that each voxel contains the same number $2N$ of points, by traversing each point in the global map and searching for $2N$ nearest neighbors, recorded as $\breve{g}_{k-N}, \cdots, \breve{g}_k, \cdots, \breve{g}_{k+N}$. This voxel division method is extremely fast and can preserve all points with more details while mitigating the influence of the outliers. After calculating the weighted centroid $p_k$ (Equation 13) of each elastic voxel, we obtain the smoothed global map $\breve{G}' = \{\breve{g}'_1, \breve{g}'_2, \breve{g}'_3, \cdots, \breve{g}', \cdots, \breve{g}'_K\}$ (Equation 14) which is the aggregation of all centroids, where the WEVG operation is represented by notation $\widehat{w}$.

$$p_k = \left(\sum_{l=k-N}^{k+N} \rho_l \breve{g}_l\right) \bigg/ \left(\sum_{l=k-N}^{k+N} \rho_l\right) \tag{13}$$

$$\breve{G}' = \{\breve{g}'_1, \breve{g}'_2, \breve{g}'_3, \cdots, \breve{g}'_k, \cdots, \breve{g}'_K\} = \{p_1, p_2, p_3, \cdots, p_k, \cdots, p_K\} = \widehat{w}[\breve{G}] \tag{14}$$

3.2.4 Cross Section Extraction

The cross sections are vitally important for the development and dynamics of channelized debris flow. The typical cross sections can also be used to back-calculate the peak flow discharge of debris flow (Whipple, 1997; Berti et al., 1999). Hence, we have developed an algorithm to extract cross sections of the channel to represent the channel structure more intuitively. Cross section extraction consists of two parts: cutting and reconstruction.

The channel resembles a meandering curve, so the ideal cross section cutting plane is the normal plane of each point in the curve. It is feasible to extract the channel cross sections due to the grouping and sorting of the global map in Section 3.2.2. Each single unit $\breve{U}_j$ can be regarded as a cross section element. Moreover, the smoothed, densified, projected, and corrected trajectory $\breve{T}''$ produced in coordinate modification can also reflect the direction of the channel and guide us to determine the position and normal direction of cross sections. Because the morphology point cloud is sparse, it requires fusing several adjacent elements into a new point cloud $\breve{U}'_j = \{\breve{u}'_{j1}, \breve{u}'_{j2}, \breve{u}'_{j3}, \cdots, \breve{u}'_{jm}, \cdots, \breve{u}'_{jM}\}$. Then $\breve{U}'_j$ is reprojected onto one single plane to obtain a 2D cross section $\breve{U}''_j = \{\breve{u}''_{j1}, \breve{u}''_{j2}, \breve{u}''_{j3}, \cdots, \breve{u}''_{jm}, \cdots, \breve{u}''_{jM}\}$. When we combine the nearest $2a$ units of $\breve{U}_j$ and obtain the new cloud $\breve{U}'_j$, we can connect it to the division index trajectory points, $\breve{t}''_{j-a}, \cdots, \breve{t}''_j, \cdots, \breve{t}''_{j+a}$. The trajectory point $\breve{t}''_j$ locates on the projection plane. The normal of the projection plane is $\breve{t}''_{j+a} - \breve{t}''_{j-a}$. Then we can determine all points of $\breve{U}''_j$ by establishing and solving a parametric equation (Equation 15), where a new parameter $\alpha$ (Equation 16) was introduced here.

$$\breve{u}''_{jm} = \breve{u}'_{jm} - \alpha(\breve{t}''_{j+a} - \breve{t}''_{j-a}) \tag{15}$$

$$\alpha = (\breve{u}'_{jm} - \breve{t}''_j)(\breve{t}''_{j+a} - \breve{t}''_{j-a})^T / \|\breve{t}''_{j+a} - \breve{t}''_{j-a}\| \tag{16}$$

However, the new cross-section $\breve{U}''_j$ is not a curve with a clear boundary but accompanied by many fluctuating noises which is not qualified for further study. Furthermore, there are many blanks caused by water surface refraction and rock obscuration in the global map, which might make cross-section discontinuous. This is because during data acquisition, the LIDAR scanning may not cover certain sections of the channel, either because they are unreachable or obscured by the harsh channel environment. These deficiencies impede analyzing of channel structure for subsequent application. Therefore, we smooth cross section point clouds again with our new smoothing method, WEVG, and then densify it by linear interpolation to gain the final cross section $\breve{U}'''_j$ (Equation 17). Before the linear interpolation, we sort the cross-section in the order of height for wall point and width for ground point.

$$\breve{U}'''_j = \hat{L}(sort|\widehat{w}[\breve{U}''_j|]) \tag{17}$$

## 4 Outputs of AscDAMs

The morphological map generated by AscDAMs contains detailed channel morphology data. Section 4.1 explains the selection criteria and the overall characteristics of the optimal maps. Detailed channel morphology including the typical cross section and basal erosion will be introduced in Section 4.2. The delineation of channel deposits distinction, volume estimation, and change monitoring is presented in Section 4.3.

**4.1 The Global Map of the Channel**

Mapping results calculated with different parameters of down-sampling and loop closure will be various. Finding appropriate parameters is a prerequisite for successful and accurate mapping. Among these parameters, the down-sampling rate is an uncertainty factor while the remaining factors either possess predetermined values or exert negligible influence on the mapping accuracy.

After obtaining the trajectory and global map, we projected them onto the DOM of ZY03 satellite with their elevation data, as shown in Figure 5, where the relative elevation in the graph is based on the location of the start point of the trajectory (Figure 5a). The optimal point cloud global map is shown in Figure 5b. Detailed channel morphology represented by different typical cross sections (TCS) (Figure 5b) is selected for further demonstration.

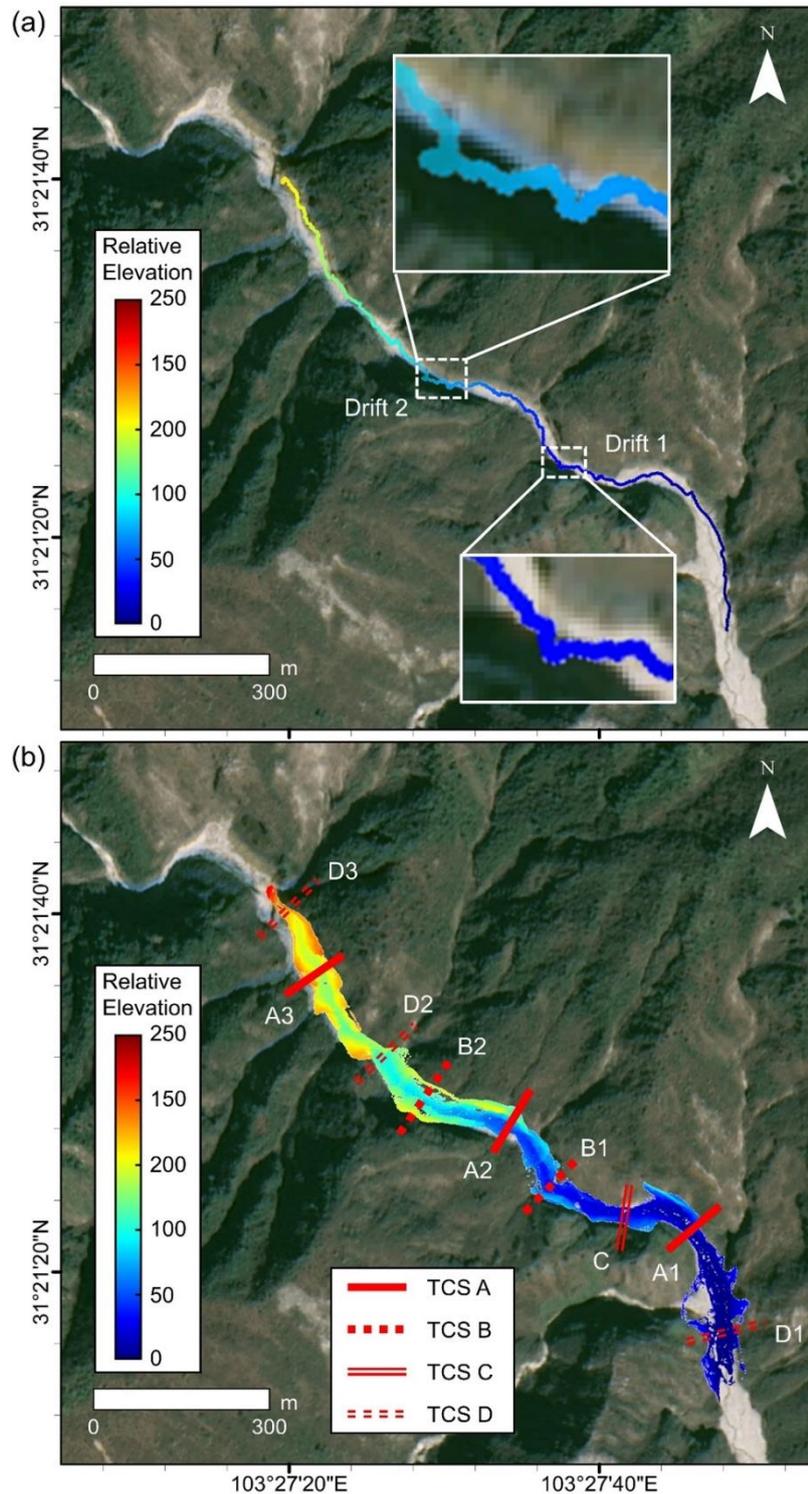

*Figure 5. (a) The trajectory and (b) the global map, TCS A~D are typical cross sections of the channel.*

The uneven trajectory (Figure 5a) shows the ruggedness of channel interior and the difficulty of employing original LIOSAM to such harsh environment. It is clear to see that the trajectory we calculated and channel area on DOM are highly matching. The global map (Figure 5b) presents the same good fitting. The structure, outline, and confluences of the channel point cloud match well with the satellite image obviously, although there is a slight offset at the upstream end. We can also see that the height of the cliff captured in the map is tens of meters (Figure 5b). The maximum elevation difference of the channel bed within the trajectory range is about 160 m. Two drifts on the trajectory seem incongruent with the DOM and will be discussed in Section 5.1.3.

The DEM with a horizontal resolution and elevation accuracy of 0.1 m was derived from the global map and is shown in Figure 6a. It should be noted that the DEM generated from AscDAMs was based on the filtered point cloud global map. Because of the existence of basal erosion and overhanging cliff, there could be several elevation values at the same raster. Hence, the redundant elevation values should be removed to avoid unnecessary errors although this filtering process will lead to a decrease in channel morphology information. The DEM obtained by AscDAMs can be integrated with DEM generated by satellites and/or UAVs which can be used in numerical simulation of debris flows.

The colored information is also integrated with the global map by running RGB-SLAM, enhancing the comprehension of the channel interior information. The channel map is predominantly gray-brown after rendering because the data were collected in February when most of the vegetation did not sprout. In the map, unconsolidated sediment is grey while withered grass is brownish (Figure 6b). The long-standing stable deposits are brown (Figure 6c). There are many places in the channel where the color is dark green. These sections are usually full of fresh plants (Figure 6c), while bedrock in the shadow might exhibit the same color (Figure 6d). The colored topographic map can be utilized for determining the deposit stability and vegetation restoration.

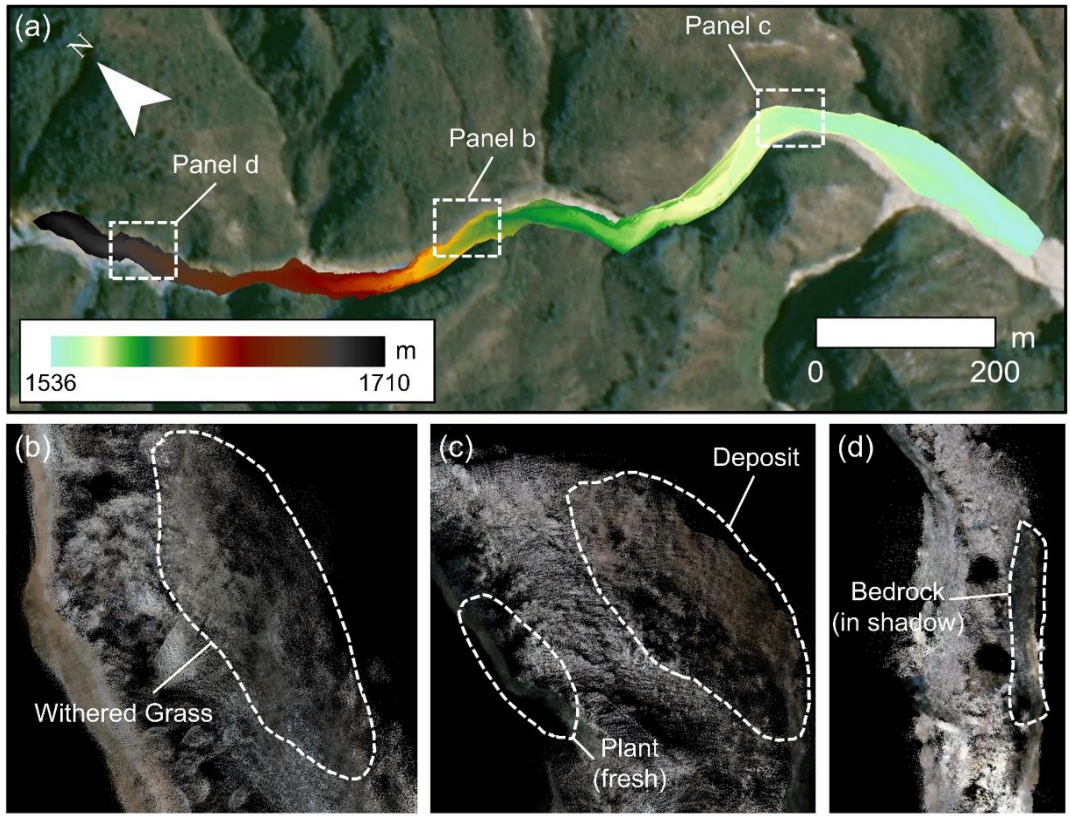

*Figure 6. The channel DEM and colored map. (a) The AscDAMs' DEM of the channel. (b)Withered Grass. (c)Deposit and fresh plant. (d) Bedrock in shadow.*

## 4.2 Detailed Channel Morphology

4.2.1 Typical Cross section

Although the channel morphology varies throughout its entirety, the geomorphic features within the channel can be reflected by different typical cross sections as shown in Figure 7, such as artificial facilities (TCS A1), rocks (TCS A2), and pools (TCS A3). Man-made facilities like the small dam are the easiest to identify due to their regular shape while the natural landscape need more attentiveness. By cutting out and densifying cross sections, the various shapes of the channel are more intuitive. Typically, the cross sections of the channel are V-shaped or U-shaped, featuring a narrow bottom surface, wide top and steep side walls.

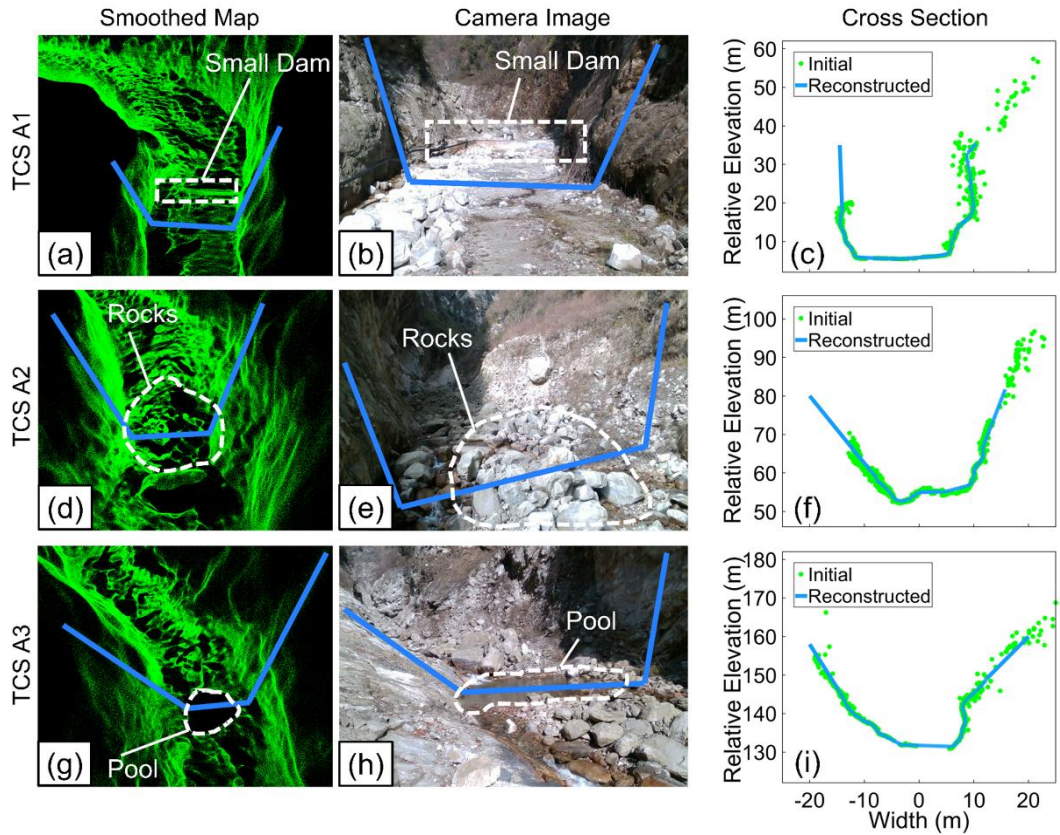

*Figure 7. Inside view of TCS A1, A2, A3 in the smoothed map (a) (d) (g), camera images (b) (e) (h), and cross sections before and after reconstruction (c) (f) (i).*

At the channel bed, the ground is uneven with deposit materials having a wide grain size distribution from silt/clay to cobbles and huge rocks. Since the water bodies will reflect and refract the LIDAR signal, there are some missing parts in the point cloud. For distinguishing blanks, narrow, continuous, and low-lying areas on the ground are mostly streams, while large horizontal gaps are mostly pools. The other types of blanks are mostly due to missing scans. On both sides of the channel, the sleek channel boundaries correspond to clear and smooth point clusters, while vegetations and loose material correspond to fuzzy and rough clouds. The distinct and detailed morphology at the bottom of the channel, including the channel bed and side walls, which is significant to the development of debris flows, cannot be obtained from existing satellite images.

### 4.2.2 Erosion Pattern

Debris flow erosion consists of two parts, bed entrainment and bank erosion (Stock

& Dietrich, 2006; Berger et al., 2011a). Bank erosion is important for channelized debris flow, especially when debris flow changes its direction. Satellite images could not provide sufficient channel interior information for the study of erosion. However, the present study successfully addressed this bottleneck technique question using the global map and extracted cross sections. As an illustration, AscDAMs results distinctly reveal two areas of bank erosion as shown in Figure 8. The cliffs lean towards the channel interior because their middle and lower parts have been eroded heavily. These erosions can be caused by the impingement of debris flows and/or other frictions. From the perspective of the DOM, shadow on these two segments is heavy. It indicates that the cliffs are towering and steep so that the environment inside the channel is difficult to detect. Thus, the AscDAMs hold a unique advantage in detecting lateral bank erosion and providing supplementary information that cannot be observed from satellite images and UAVs.

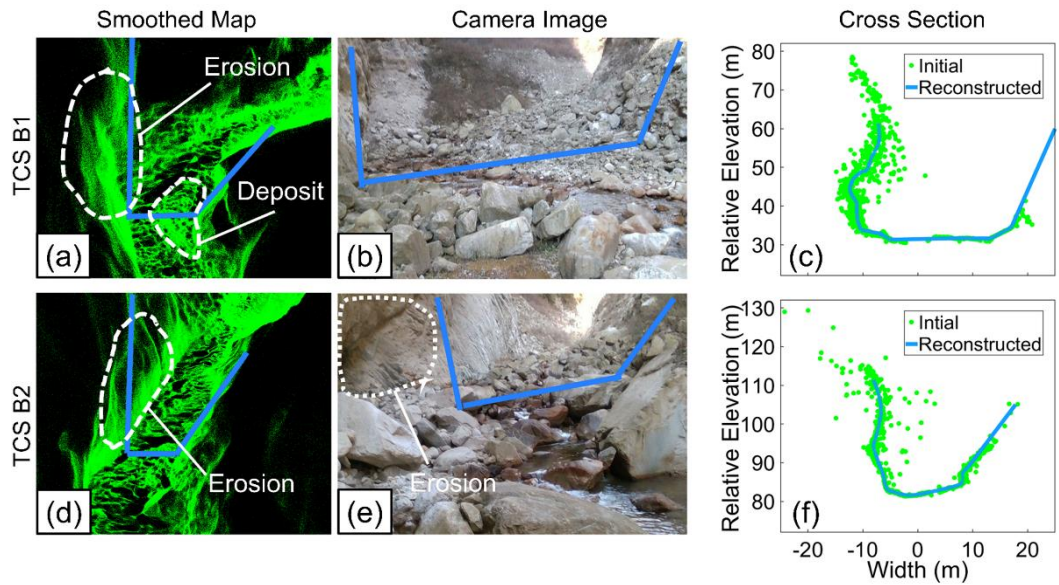

Figure 8. The erosion pattern in TCS B1 (a-c), and B2 (d-f) with large trajectory drift in Figure 5a.

## 4.3 Channel Deposit Distinction and Monitoring

4.3.1　Channel Deposit Distinction and Volume Estimation

Deposit in the channel can be identified directly by the global map and cross sections. At the channel bed, there are scattered stones in most places, which are either uneven on the surface, such as TCS A2 (Figure 7), or inclined to accumulate at the foot of the sidewall, such as TCS B1 (Figure 8). Besides, individual large deposits can also be clearly recorded by the global map. For instance, at TCS C (Figure 9), a body of deposit can be identified of about 7 m high, occupying about half of the channel bed width, next to the north channel wall. The erosion pattern on this deposit can be clearly identified that the part outside of the bend was severely eroded. Furthermore, the volume of deposits can be estimated with the coordinate data of the global map and cross sections. The deposit in TCS C, for example, can be approximated as a combination of regular polyhedrons as shown in (Figure 10a). Assuming the bedrock surface behind the deposit is flat and regular, we can divide the deposit body of TCS C into a quadrilateral prism with a trapezoidal base and a triangular prism with a trapezoidal side. The volumes of the two prisms are 661 $m^3$ and 504 $m^3$ respectively. Hence, the total volume is estimated to be 1165 $m^3$. Using the same method, the deposit volume in TCS A2 (Figure 10b) and TCS B1 (Figure 10c) is evaluated to be 540 $m^3$ and 595 $m^3$ respectively. Note that this is a rough estimation as the deposit boundary inside is not known. If the channel morphology is detected by AscDAMs before and after the debris flow event, the accuracy of estimating volume change could be significantly improved.

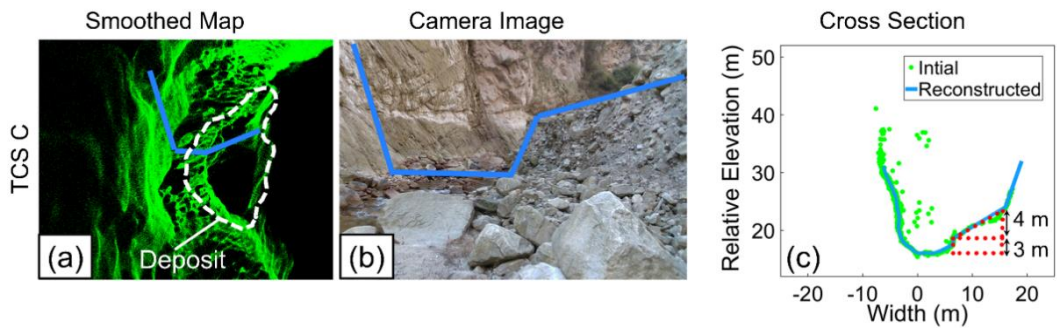

Figure 9. The deposit in TCS C and volume estimation.

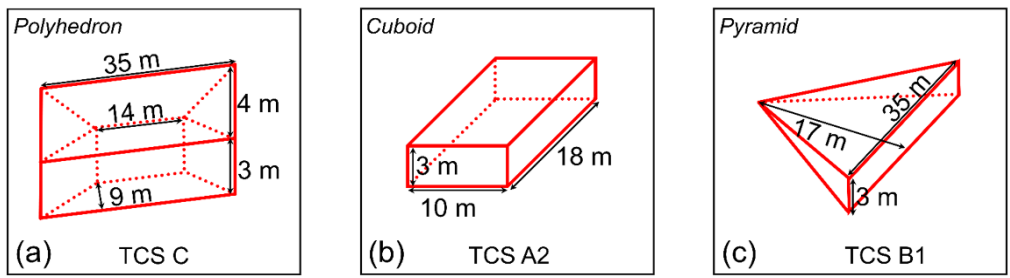

Figure 10. Deposit volume estimation.

A proper estimation of possible source material of debris flow is vital to estimate the volume and destructive power of debris flow which are significant to hazard mitigation. However, the deposits are difficult to identify and evaluate from DOM or DEM of satellite, especially in narrow alpine channels. Satisfactorily, it can be identified intuitively and analyzed quantitatively on the global map now, which demonstrates the technological advantage of AscDAMs.

4.3.2 Deposit Change Monitoring

Regular detection of channel morphology by AscDAMs makes it possible to study the long-term spatial and temporal evolution of channel deposits. In the current study, two field investigations are carried out in February and November of 2023 which span an entire rainy season. A new dataset was obtained by AscDAMs after a whole rainy season. By comparing the two different data, it is possible to monitor the migration of loose channel deposits during the rainy season. The results of AscDAMs show that the deposits inside the channel almost have no significant change during this period, except for a collapse of the deposit at TCS D1 as shown in Figure 11. The deposit was like a triangular prism tightly against the channel wall last time. After collapsing, it resembles a lying pyramid, with negligible change in volume. The quantitative analysis of changes in channel deposits provides valuable insights for studying debris flow risk in the study area. This aspect holds significant importance for hazard mitigation efforts.

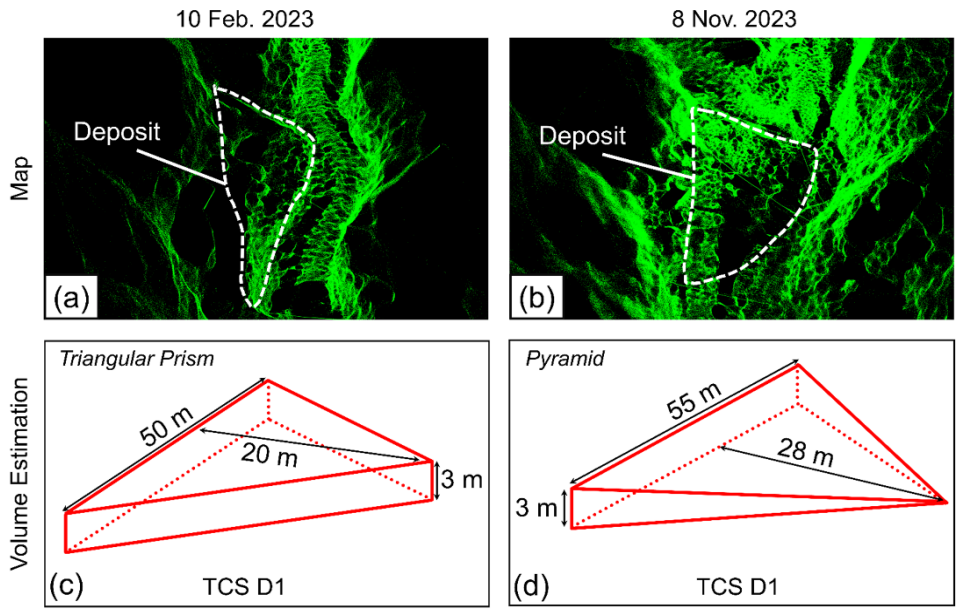

*Figure 11. The deposit change in TCS D1.*

## 5 Discussion

### 5.1 The Quality of Smoothing and Extraction

The quality of the point cloud smoothing can be exemplified by TCS D1, D2, and D3 (Figure 12). The position of TCS D1 is a confluence point of the tributary to the main channel of Chutou Gully (Figure 5b). A small alluvial fan was formed at the outlet of the tributary. The surface of the alluvial fan and rocks in the smoothed images is clearer than that in the initial map. The smoothed point cloud of an artificial wire is clearer and more distinguishable. TCS D2 is another intersection. A 300-$m^3$ deposit's outline in front of a tributary outlet is more distinguishable after smoothing. There is a pool at TCS D3. The boundary of the pool is more distinct after smoothing and there are significantly fewer noises on the adjacent rock wall. Compared with the point cloud images before smoothing, the visual effect after smoothing is significantly improved with clearer material boundaries and a lighter blurriness. It makes observing the landforms inside the channel much more convenient and accurate.

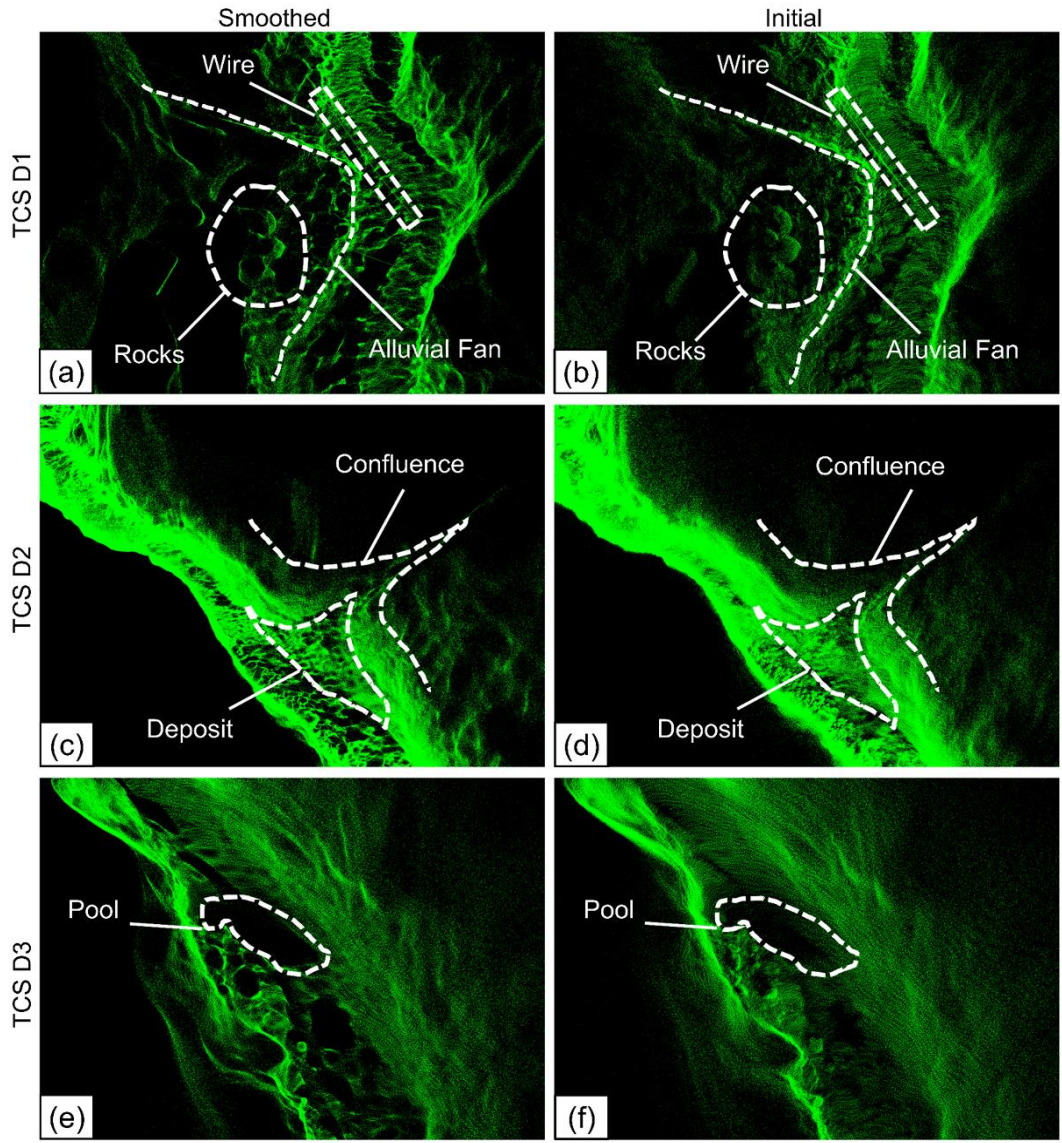

*Figure 12. Point clouds of TCS D1 (a, b), D2 (c, d), D3 (e, f) before and after smoothing.*

The quality of cross-section extraction is also satisfactory. In cross section extraction, both images before and after reconstruction are preserved (Figure 7, Figure 8, Figure 9). Point clouds of cross-sections before reconstruction are sparse and cluttered while the reconstructed are dense and ordered. Even if reconstructed cross-section has some offset on occasion (Figure 9c), it does not affect its description of the channel profile. For comparison, we created cross-sections of TCS A1 and TCS B1 with the data obtained by laser range finder (LRF) from field investigation and the DEM from ZY03 stereo images, as shown in Figure 13. Compared with the LRF and ZY03's DEM data, the reconstructed cross-sections and AscDAMs' DEM are more accurate with more channel structure information. Both ZY03's DEM and LRF data lose channel details. Moreover, the errors of ZY03's DEM are apparently large and cannot reflect lateral erosion of the channel inner wall and erosion of deposits. In general, the AscDAMs' results are superior to the common-used data from existing methods. Their accuracy is much higher than that of the prevailing satellite-derived DEM. Such high-quality cross sections and DEM can be further used for simulating and even predicting debris flow. This precision is crucial for a comprehensive understanding of debris flow mechanisms.

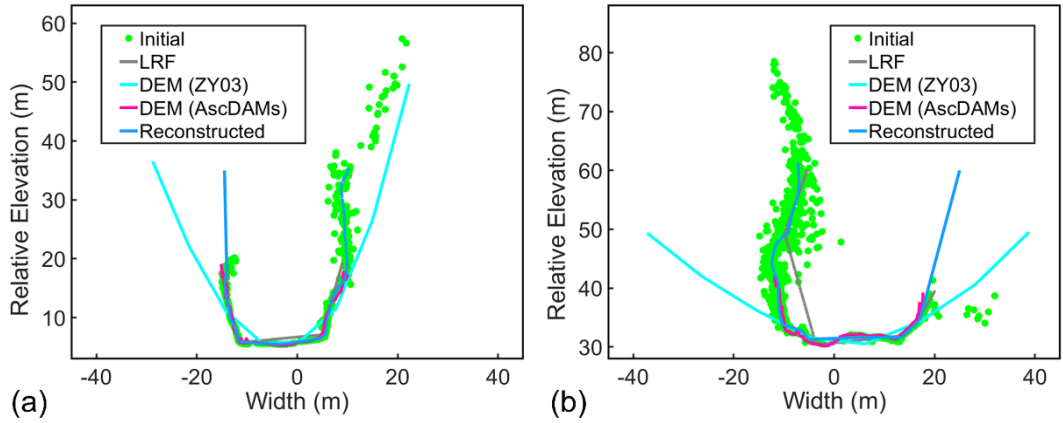

*Figure 13. Cross-sections from ZY03's DEM, LRF, AscDAMs' DEM, and global map before and after reconstruction: (a) TCS A1. (b) TCS B1.*

**5.2 The Accuracy of AscDAMs**

5.2.1 The Elevation Error of AscDAMs

We analyze the elevation error by comparing the height data of the initial trajectory obtained by LIOSAM, the modified trajectory optimized by AscDAMs, barometer data, and DEM of ZY03 (Figure 14). Taking the barometer result as benchmark, the average bias and root mean square errors (RMSE) of the LIOSAM, AscDAMs, and DEM trajectories are estimated through polynomial fitting of the height data (Table 3). Evidently, the DEM of ZY03 exhibits significant errors within the research area, featuring abnormal sudden rises and drops of elevation that deviate from the field investigations. The fluctuation of elevations of ZY03 can reach 100 m from Figure 14. It is found that the AscDAMs trajectory has the most precise elevation data with an average bias of -0.48 m and a root mean square error of 8.65 m. This affirms the effectiveness of the proposed deviation correction algorithm. The satellite-derived elevation, namely DEM of ZY03 is incorrect, with many unreal fluctuations along the channel as shown in Figure 14, which is due to the inability of satellite to detect the channel bed where it is narrow and in shadow (Cao et al., 2021).

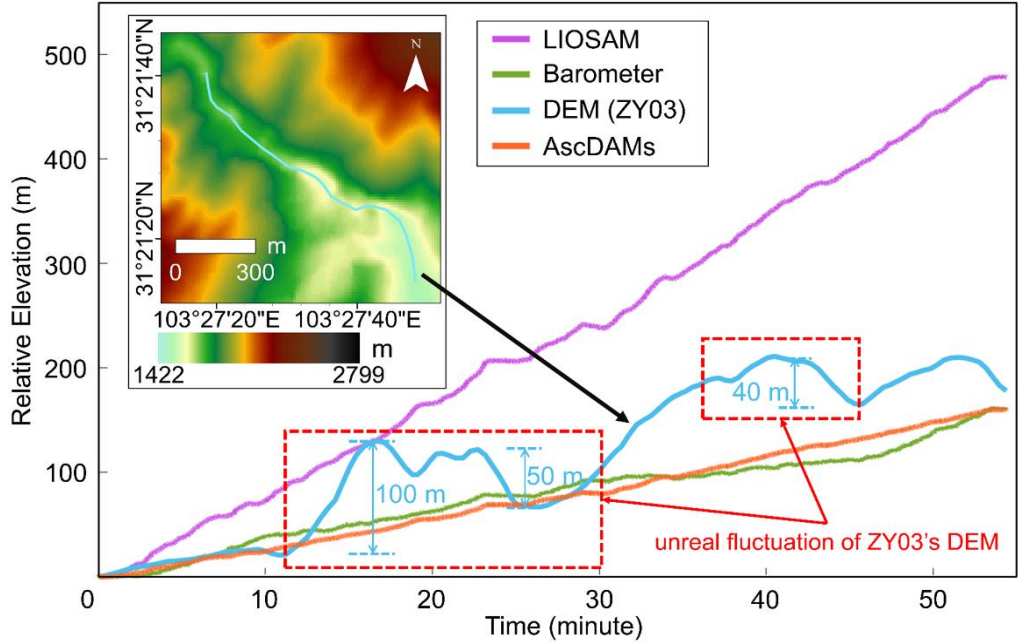

*Figure 14. The height data of the initial trajectory obtained by LIOSAM, barometer, DEM of ZY03 on 14 January 2021, and the modified trajectory optimized by AscDAMs.*

*Table 3. The average elevation bias and RMSE of the LIOSAM, AscDAMs, and DEM trajectories.*

|  | LIOSAM | AscDAMs | DEM |
|---|---|---|---|
| Average bias (m) | 152.76 | -0.48 | 39.25 |
| RMSE (m) | 182.43 | 8.65 | 46.91 |

5.2.2 The Horizontal Error of AscDAMs

The bias, displacement error and distance error between the end point of the trajectories and the referred end point are calculated respectively based on the DOM (Figure 15), as shown in Table 4. The reference displacement, which is the straight distance between the referred start and end points in DOM, is 1090 m. The reference distance, which is the estimated length of the channel from the referred start point and referred end point, is 1190 m. The horizontal bias of AscDAMs is 28.66 m, resulting in the displacement error and distance error only 2.63% and 2.43%, respectively. As a comparison, the initial maps of LIOSAM have a large offset of 182.68 m. This shows that the proposed deviation correction algorithm in AscDAMs can significantly improve the horizontal accuracy from the original LIOSAM result.

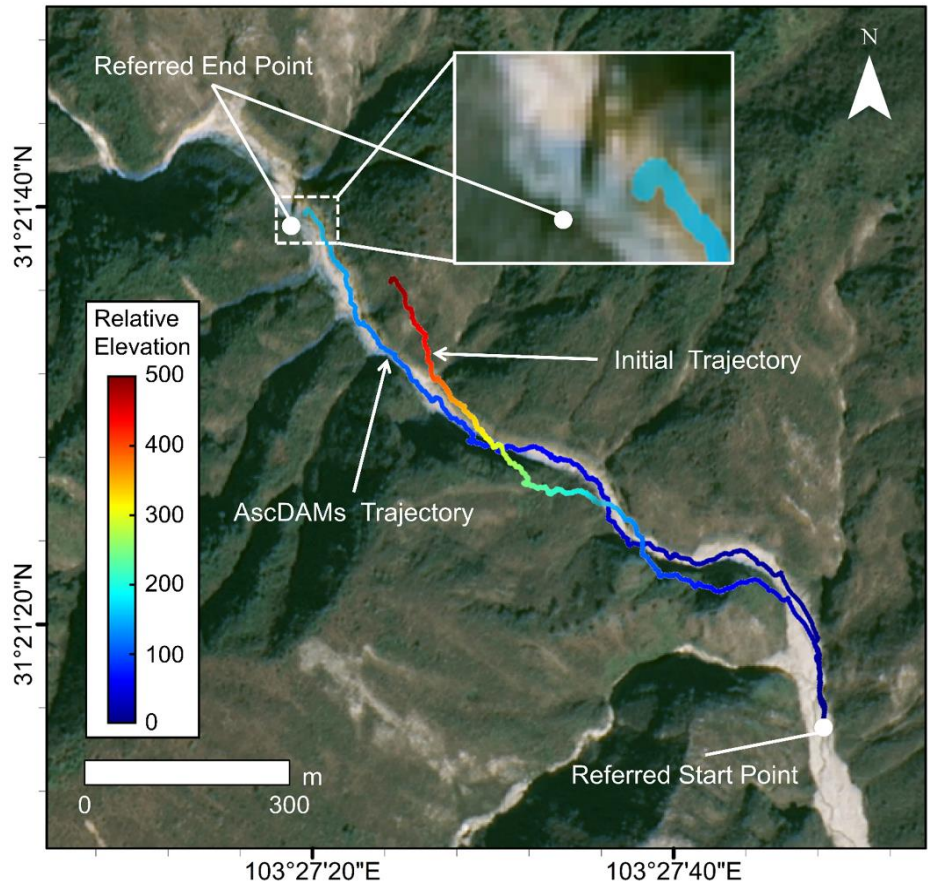

*Figure 15. The trajectories of AscDAMs and LIOSAM.*

*Table 4. Horizontal bias of AscDAMs and LIOSAM.*

| Algorithm | End point bias (m) | Displacement error (%) | Distance error (%) |
|---|---|---|---|
| AscDAMs | 28.66 | 2.63 | 2.43 |
| LIOSAM | 182.68 | 16.76 | 15.35 |

5.2.3  Drifts and Bank Erosion

As mentioned above, there are two irregular deviations of trajectory, Drift 1 and Drift 2 (Figure 5a). However, these two drifts are not introduced by the accumulation of systematic error. By comparing the locations of drifts and selected bank erosion areas, it is easy to find that the drifts correspond to the bank erosion areas. The obscuration by overhanging cliffs prevents the DOM from capturing accurate channel morphology. Consequently, the real channel morphology could not accurately be depicted on the DOM of ZY03, contributing to the confusion of "drifts". This shows the superiority of AscDAMs to capture details of channel interior structure over satellite images.

**5.3 The Universality and Usability of AscDAMs**

AscDAMs succeed in promoting SLAM to complex gully environments to obtain high-resolution topographic maps with full characteristics of debris flow channels. In the previous study, SLAM has been tested in small-scale hillslope gully with gentle terrain (Kinsey-Henderson et al., 2021) and broad hillsides with stable operating conditions for SLAM (Marotta et al., 2021). In the current study, the research area was selected as the most challenging environment for the AscDAMs. The DOM-aided deviation correction algorithm effectively minimized the error and greatly enhanced the accuracy of the SLAM results. The point cloud smoothing algorithm mitigated the effects of sensors swaying and rocking, leading to a substantial improvement in the quality of the final point cloud map. By utilizing the cross section extraction algorithm, the channel morphology could be quantitatively assessed. The location and volume of channel deposits can be precisely quantified. The change of each deposit could also be accurately detected by comparing two different AscDAMs maps. The successful implementation of AscDAMs in such a challenging environment implies that for other channels in the Wenchuan earthquake region or less demanding scenarios with more typical features for the SLAM algorithms to compute, favorable results can also be achieved.

AscDAMs is a multi-sensor fusion system. Compared to existing channel detection technologies, AscDAMs is easy, economical, and effective. It just assembles with only three core components, LIDAR, IMU, minicomputer. As complementary to the current field investigation methods, our system can record the channel morphology quantitatively and automatically while traveling along the channel without manual supervision or intervention. AscDAMs does not need complex or elaborate route planning in advance. Furthermore, without any special requirement of hardware, an ordinary computer is sufficient for this calculation, and the computing time is equivalent to the data collection duration. If higher accuracy is pursued, the global map can be further optimized with input from magnetic sensors, altimeters, and other sensors. AscDAMs can utilize different sensors according to specific environments while the underlying algorithm's logic and processing flow remain unchanged.

### 5.4 The Limitations of AscDAMs

Although AscDAMs offer advantages in terms of accuracy, novelty, and efficiency, it does exhibit certain limitations to some extent.

(1) The final global map will become sparse visually after smoothing for point cloud resampling (Liu & Zhong, 2014). This is the common defect of point cloud noise filtering and smoothing algorithms.

(2) The color contrast of the colored map is insufficient. Vegetation changes with season. In this study, we only tested AscDAMs before Spring when plants had not yet recovered. More tests can be implemented in different seasons. In addition, the complexity of the light affects the color recognition of the camera.

(3) The equipment is applicable for areas within the reach of manpower. In some cases, channels might be blocked for walking. AscDAMs can be combined with more topography-adaptive carriers in the future.

### 6  Conclusion and Perspective

Obtaining the high-resolution, accurate topographic channel map is the common key challenge for channelized debris flow research. At present, wide-used satellite images, UAV-based mapping, and other existing technologies cannot satisfy the requirements of accuracy and efficiency in observing channel interior conditions in mountainous long-deep gullies. SLAM is an emerging 3D mapping tech and has been applied across different platforms for numerous scenarios of topographic mapping. However, state-of-art SLAM mapping results contain large drift and abundant noise induced by the extremely rugged long-deep channel environment. Aiming to solve these problems, we proposed AscDAMs with a set of new algorithms including deviation correction, point cloud smoothing, cross section reconstruction to process the original

SLAM results. In addition, a map coloring algorithm is developed to supplement more information to the map. A frequent debris flow gully named Chutou Gully in Wenchuan Earthquake region was selected as the research area. The outputs of AscDAMs demonstrate high accuracy and low noise, enabling precise detection of channel morphology and erosion pattern. Furthermore, AscDAMs facilitates the quantitative evaluation of channel deposits, encompassing spatial distribution, volume estimation, and change monitoring. These capabilities serve to compensate for the limitations of existing channel detection methods.

In conclusion, we succeed in obtaining the high-resolution full-character morphological map of debris flow gully in extremely harsh environments by utilizing the proposed novel advanced detection technology. The high-resolution DEM and channel cross-section profile derived from AscDAMs can be used directly in the numerical simulation of debris flow. AscDAMs map facilitates the straightforward identification and quantitative analysis of channel deposits, which makes it possible to back-calculate the sediment transportation process inside the channel by comparing two different AscDAMs maps. The color information is integrated and explicitly displayed on the AscDAMs map, enabling ones to study the vegetation recovery inside the channel. As a crucial supplement to existing channel morphology detection methods, AscDAMs works well in the complex channel environments. It provides the important but currently absent channel interior details, which is promising to promote deep understanding of debris flow mechanisms and post-seismic long-term evolution, and support precise hazard/risk assessment and mitigation, although it can be further improved in systemic error correction.

**Acknowledgments**

The authors greatly acknowledge the financial support from the Science and Technology Development Fund, Macao SAR (File Nos. 0083/2020/A2 and SKL-IOTSC(UM)-2021-2023), and the National Natural Science Foundation of China (Nos. 42007245).

# References

Bailey, T., & Durrant-Whyte, H., 2006. Simultaneous localization and mapping (SLAM): Part II. Ieee Robotics & Automation Magazine, 13(3), 108-117. https://doi.org/10.1109/Mra.2006.1678144

Barros, A. M., Michel, M., Moline, Y., Corre, G., & Carrel, F., 2022. A Comprehensive Survey of Visual SLAM Algorithms. Robotics, 11(1). https://doi.org/10.3390/robotics11010024

Berger, C., McArdell, B. W., & Schlunegger, F., 2011a. Direct measurement of channel erosion by debris flows, Illgraben, Switzerland. Journal of Geophysical Research-Earth Surface, 116. https://doi.org/10.1029/2010jf001722

Berger, C., McArdell, B. W., & Schlunegger, F., 2011b. Sediment transfer patterns at the Illgraben catchment, Switzerland: Implications for the time scales of debris flow activities. Geomorphology, 125(3), 421-432. https://doi.org/10.1016/j.geomorph.2010.10.019

Berti, M., Genevois, R., Simoni, A., & Tecca, P. R., 1999. Field observations of a debris flow event in the Dolomites. Geomorphology, 29(3-4), 265-274. https://doi.org/10.1016/S0169-555x(99)00018-5

Blasone, G., Cavalli, M., Marchi, L., & Cazorzi, F., 2014. Monitoring sediment source areas in a debris-flow catchment using terrestrial laser scanning. Catena, 123, 23-36. https://doi.org/10.1016/j.catena.2014.07.001

Bonneau, D. A., Hutchinson, D. J., McDougall, S., DiFrancesco, P. M., & Evans, T., 2022. Debris-Flow Channel Headwater Dynamics: Examining Channel Recharge Cycles With Terrestrial Laser Scanning. Frontiers in Earth Science, 10. https://doi.org/10.3389/feart.2022.883259

Bradley, D., & Roth, G., 2007. Adaptive thresholding using the integral image. Journal of graphics tools, 12(2), 13-21. https://doi.org/10.1080/2151237X.2007.10129236

Cadena, C., Carlone, L., Carrillo, H., Latif, Y., Scaramuzza, D., Neira, J., Reid, I., & Leonard, J. J., 2016. Past, Present, and Future of Simultaneous Localization and Mapping: Toward the Robust-Perception Age. Ieee Transactions on Robotics, 32(6), 1309-1332. https://doi.org/10.1109/Tro.2016.2624754

Caduff, R., Schlunegger, F., Kos, A., & Wiesmann, A., 2015. A review of terrestrial radar interferometry for measuring surface change in the geosciences. Earth Surface Processes and Landforms, 40(2), 208-228. https://doi.org/10.1002/esp.3656

Cao, C., Zhang, W., Chen, J. P., Shan, B., Song, S. Y., & Zhan, J. W., 2021. Quantitative estimation of debris flow source materials by integrating multi-source data: A case study. Engineering Geology, 291. https://doi.org/10.1016/j.enggeo.2021.106222

Chen, H. X., & Zhang, L. M., 2015. EDDA 1.0: integrated simulation of debris flow erosion, deposition and property changes. Geoscientific Model Development, 8(3), 829-844. https://doi.org/10.5194/gmd-8-829-2015

Chen, M., Tang, C., Xiong, J., Chang, M., & Li, N., 2024. Spatio-temporal mapping and long-term evolution of debris flow activity after a high magnitude earthquake. Catena, 236, 107716. https://doi.org/10.1016/j.catena.2023.107716

Cucchiaro, S., Cavalli, M., Vericat, D., Crema, S., Llena, M., Beinat, A., Marchi, L., & Cazorzi, F., 2019. Geomorphic effectiveness of check dams in a debris-flow catchment using multi-temporal topographic surveys. Catena, 174, 73-83. https://doi.org/10.1016/j.catena.2018.11.004

Durrant-Whyte, H., & Bailey, T., 2006. Simultaneous localization and mapping: Part I. Ieee Robotics &

Automation Magazine, 13(2), 99-108. https://doi.org/10.1109/Mra.2006.1638022

Fan, X. M., Juang, C. H., Wasowski, J., Huang, R. Q., Xu, Q., Scaringi, G., van Westen, C. J., & Havenith, H. B., 2018. What we have learned from the 2008 Wenchuan Earthquake and its aftermath: A decade of research and challenges. Engineering Geology, 241, 25-32. https://doi.org/10.1016/j.enggeo.2018.05.004

Fan, X. M., Scaringi, G., Korup, O., West, A. J., van Westen, C. J., Tanyas, H., Hovius, N., Hales, T. C., Jibson, R. W., Allstadt, K. E., Zhang, L. M., Evans, S. G., Xu, C., Li, G., Pei, X. J., Xu, Q., & Huang, R. Q., 2019. Earthquake-Induced Chains of Geologic Hazards: Patterns, Mechanisms, and Impacts. Reviews of Geophysics, 57(2), 421-503. https://doi.org/10.1029/2018rg000626

GitHub, Camera_lidar_fusion. https://github.com/KAI-yq/camera_lidar_fusion, 2021 (accessed 25 May 2023)

Guo, X. J., Cui, P., Li, Y., Zou, Q., & Kong, Y. D., 2016. The formation and development of debris flows in large watersheds after the 2008 Wenchuan Earthquake. Landslides, 13(1), 25-37. https://doi.org/10.1007/s10346-014-0541-6

Hu, T., & Huang, R. Q., 2017. A catastrophic debris flow in the Wenchuan Earthquake area, July 2013: characteristics, formation, and risk reduction. Journal of Mountain Science, 14(1), 15-30. https://doi.org/10.1007/s11629-016-3965-8

Huang, G. H., Lv, G. S., Zhang, S., Huang, D. L., Zhao, L. H., Ni, X. Q., Liu, H. W., Lv, J. H., & Liu, C. D., 2022. Numerical analysis of debris flows along the Sichuan-Tibet railway based on an improved 3D sphere DDA model and UAV-based photogrammetry. Engineering Geology, 305. https://doi.org/10.1016/j.enggeo.2022.106722

Imaizumi, F., Masui, T., Yokota, Y., Tsunetaka, H., Hayakawa, Y. S., & Hotta, N., 2019. Initiation and runout characteristics of debris flow surges in Ohya landslide scar, Japan. Geomorphology, 339, 58-69. https://doi.org/10.1016/j.geomorph.2019.04.026

Kinsey-Henderson, A., Hawdon, A., Bartley, R., Wilkinson, S. N., & Lowe, T., 2021. Applying a Hand-Held Laser Scanner to Monitoring Gully Erosion: Workflow and Evaluation. Remote Sensing, 13(19). https://doi.org/10.3390/rs13194004

Kukko, A., Kaijaluoto, R., Kaartinen, H., Lehtola, V. V., Jaakkola, A., & Hyyppa, J., 2017. Graph SLAM correction for single scanner MLS forest data under boreal forest canopy. ISPRS Journal of Photogrammetry and Remote Sensing, 132, 199-209. https://doi.org/10.1016/j.isprsjprs.2017.09.006

Li, J. P., Wu, W. T., Yang, B. S., Zou, X. H., Yang, Y. D., Zhao, X., & Dong, Z., 2023. WHU-Helmet: A Helmet-Based Multisensor SLAM Dataset for the Evaluation of Real-Time 3-D Mapping in Large-Scale GNSS-Denied Environments. Ieee Transactions on Geoscience and Remote Sensing, 61. https://doi.org/10.1109/Tgrs.2023.3275307

Li, Z. H., Chen, J. P., Tan, C., Zhou, X., Li, Y. C., & Han, M. X., 2021. Debris flow susceptibility assessment based on topo-hydrological factors at different unit scales: a case study of Mentougou district, Beijing. Environmental Earth Sciences, 80(9). https://doi.org/10.1007/s12665-021-09665-9

Liang, W. J., Zhuang, D. F., Jiang, D., Pan, J. J., & Ren, H. Y., 2012. Assessment of debris flow hazards using a Bayesian Network. Geomorphology, 171, 94-100. https://doi.org/10.1016/j.geomorph.2012.05.008

Lin, J., & Zhang, F., 2022a. R3LIVE: A Robust, Real-time, RGB-colored, LiDAR-Inertial-Visual tightly-coupled state Estimation and mapping package, in: 2022 International Conference on Robotics

and Automation, Philadelphia. https://doi.org/10.1109/ICRA46639.2022.9811935

Lin, J., & Zhang, F., 2022b. R3LIVE: A Robust, Real-time, RGB-colored, LiDAR-Inertial-Visual tightly-coupled state Estimation and mapping package., in: 2022 International Conference on Robotics and Automation, Philadelphia. https://doi.org/10.1109/ICRA46639.2022.9811935

Liu, H. H., Zhao, Y. J., Wang, L., & Liu, Y. Y., 2021. Comparison of DEM accuracies generated from different stereo pairs over a plateau mountainous area. Journal of Mountain Science, 18(6), 1580-1590. https://doi.org/10.1007/s11629-020-6274-1

Liu, Y., & Zhong, R. F., 2014. Buildings and Terrain of Urban Area Point Cloud Segmentation based on PCL, in: 35th International Symposium on Remote Sensing of Environment, Beijing. https://doi.org/10.1088/1755-1315/17/1/012238

Luo, S. Y., Xiong, J. N., Liu, S., Hu, K. H., Cheng, W. M., Liu, J., He, Y. F., Sun, H. Z., Cui, X. J., & Wang, X., 2022. New Insights into Ice Avalanche-Induced Debris Flows in Southeastern Tibet Using SAR Technology. Remote Sensing, 14(11). https://doi.org/10.3390/rs14112603

Marotta, F., Teruggi, S., Achille, C., Vassena, G. P. M., & Fassi, F., 2021. Integrated Laser Scanner Techniques to Produce High-Resolution DTM of Vegetated Territory. Remote Sensing, 13(13). https://doi.org/10.3390/rs13132504

Mergili, M., Fischer, J. T., Krenn, J., & Pudasaini, S. P., 2017. r.avaflow v1, an advanced open-source computational framework for the propagation and interaction of two-phase mass flows. Geoscientific Model Development, 10(2). https://doi.org/10.5194/gmd-10-553-2017

Meyer, N. K., Schwanghart, W., Korup, O., Romstad, B., & Etzelmuller, B., 2014. Estimating the topographic predictability of debris flows. Geomorphology, 207, 114-125. https://doi.org/10.1016/j.geomorph.2013.10.030

Morino, C., Conway, S. J., Balme, M. R., Hillier, J., Jordan, C., Saemundsson, T., & Argles, T., 2019. Debris-flow release processes investigated through the analysis of multi-temporal LiDAR datasets in north-western Iceland. Earth Surface Processes and Landforms, 44(1), 144-159. https://doi.org/10.1002/esp.4488

Mueting, A., Bookhagen, B., & Strecker, M. R., 2021. Identification of Debris-Flow Channels Using High-Resolution Topographic Data: A Case Study in the Quebrada del Toro, NW Argentina. Journal of Geophysical Research-Earth Surface, 126(12). https://doi.org/10.1029/2021JF006330

Pierzchala, M., Giguere, P., & Astrup, R., 2018. Mapping forests using an unmanned ground vehicle with 3D LiDAR and graph-SLAM. Computers and Electronics in Agriculture, 145, 217-225. https://doi.org/10.1016/j.compag.2017.12.034

Ram, P., & Sinha, K., 2019. Revisiting kd-tree for Nearest Neighbor Search, in: Proceedings of the 25th Acm Sigkdd International Conferencce on Knowledge Discovery and Data Mining, Anchorage. https://doi.org/10.1145/3292500.3330875

Remaître, A., Malet, J. P., & Maquaire, O., 2005. Morphology and sedimentology of a complex debris flow in a clay-shale basin. Earth Surface Processes and Landforms, 30(3), 339-348. https://doi.org/10.1002/esp.1161

Rusu, R. B., & Cousins, S., 2011. 3D is here: Point Cloud Library (PCL), in: 2011 IEEE International Conference on Robotics and Automation, Shanghai. https://doi.org/10.1109/ICRA.2011.5980567

Schurch, P., Densmore, A. L., Rosser, N. J., Lim, M., & McArdell, B. W., 2011. Detection of surface change in complex topography using terrestrial laser scanning: application to the Illgraben debris-flow channel. Earth Surface Processes and Landforms, 36(14), 1847-1859. https://doi.org/10.1002/esp.2206

Shan, T. X., Englot, B., Meyers, D., Wang, W., Ratti, C., & Rus, D., 2020. LIO-SAM: Tightly-coupled Lidar Inertial Odometry via Smoothing and Mapping, in: 2020 IEEE/RSJ International Conference on Intelligent Robots and Systems, Las Vega. https://doi.org/10.1109/Iros45743.2020.9341176

Shen, P., Zhang, L. M., Chen, H. X., & Fan, R. L., 2018. EDDA 2.0: integrated simulation of debris flow initiation and dynamics considering two initiation mechanisms. Geoscientific Model Development, 11(7), 2841-2856. https://doi.org/10.5194/gmd-11-2841-2018

Shen, P., Zhang, L. M., Fan, R. L., Zhu, H., & Zhang, S., 2020. Declining geohazard activity with vegetation recovery during first ten years after the 2008 Wenchuan earthquake. Geomorphology, 352. https://doi.org/10.1016/j.geomorph.2019.106989

Simoni, A., Bernard, M., Berti, M., Boreggio, M., Lanzoni, S., Stancanelli, L. M., & Gregoretti, C., 2020. Runoff-generated debris flows: Observation of initiation conditions and erosion-deposition dynamics along the channel at Cancia (eastern Italian Alps). Earth Surface Processes and Landforms, 45(14), 3556-3571. https://doi.org/10.1002/esp.4981

Stock, J. D., & Dietrich, W. E., 2006. Erosion of steepland valleys by debris flows. Geological Society of America Bulletin, 118(9-10), 1125-1148. https://doi.org/10.1130/B25902.1

Sun, Q., Zhang, L., Hu, J., Ding, X. L., Li, Z. W., & Zhu, J. J., 2015. Characterizing sudden geo-hazards in mountainous areas by D-InSAR with an enhancement of topographic error correction. Natural Hazards, 75(3), 2343-2356. https://doi.org/10.1007/s11069-014-1431-x

Tanduo, B., Martino, A., Balletti, C., & Guerra, F., 2022. New Tools for Urban Analysis: A SLAM-Based Research in Venice. Remote Sensing, 14(17). https://doi.org/10.3390/rs14174325

Tang, C., Zhu, J., Li, W. L., & Liang, J. T., 2009. Rainfall-triggered debris flows following the Wenchuan earthquake. Bulletin of Engineering Geology and the Environment, 68(2), 187-194. https://doi.org/10.1007/s10064-009-0201-6

Tang, Y. M., Guo, Z. Z., Wu, L., Hong, B., Feng, W., Su, X. H., Li, Z. G., & Zhu, Y. H., 2022. Assessing Debris Flow Risk at a Catchment Scale for an Economic Decision Based on the LiDAR DEM and Numerical Simulation. Frontiers in Earth Science, 10. https://doi.org/10.3389/feart.2022.821735

Ullman, M., Laugomer, B., Shicht, I., Langford, B., Ya'aran, S., Wachtel, I., Frumkin, A., & Davidovich, U., 2023. Formation processes and spatial patterning in a late prehistoric complex cave in northern Israel informed by SLAM-based LiDAR. Journal of Archaeological Science-Reports, 47. https://doi.org/10.1016/j.jasrep.2022.103745

Walter, F., Hodel, E., Mannerfelt, E. S., Cook, K., Dietze, M., Estermann, L., Wenner, M., Farinotti, D., Fengler, M., Hammerschmidt, L., Hansli, F., Hirschberg, J., McArdell, B., & Molnar, P., 2022. Brief communication: An autonomous UAV for catchment-wide monitoring of a debris flow torrent. Natural Hazards and Earth System Sciences, 22(12), 4011-4018. https://doi.org/10.5194/nhess-22-4011-2022

Whipple, K. X., 1997. Open-channel flow of Bingham fluids: Applications in debris-flow research. Journal of Geology, 105(2), 243-262. https://doi.org/10.1086/515916

Xiong, J., Tang, C., Gong, L. F., Chen, M., Li, N., Shi, Q. Y., Zhang, X. Z., Chang, M., & Li, M. W., 2022. How landslide sediments are transferred out of an alpine basin: Evidence from the epicentre of the Wenchuan earthquake. Catena, 208.

https://doi.org/10.1016/j.catena.2021.105781

Xu, Q., Zhang, S., Li, W. L., & van Asch, T. W. J., 2012. The 13 August 2010 catastrophic debris flows after the 2008 Wenchuan earthquake, China. Natural Hazards and Earth System Sciences, 12(1), 201-216. https://doi.org/10.5194/nhess-12-201-2012

Yang, Y., Tang, C. X., Cai, Y. H., Tang, C., Chen, M., Huang, W. L., & Liu, C., 2023. Characteristics of Debris Flow Activities at Different Scales after the Disturbance of Strong Earthquakes-A Case Study of the Wenchuan Earthquake-Affected Area. Water, 15(4). https://doi.org/10.3390/w15040698

Yang, Y., Tang, C. X., Tang, C., Chen, M., Cai, Y. H., Bu, X. H., & Liu, C., 2023. Spatial and temporal evolution of long-term debris flow activity and the dynamic influence of condition factors in the Wenchuan earthquake-affected area, Sichuan, China. Geomorphology, 435. https://doi.org/10.1016/j.geomorph.2023.108755

Ye, H. Y., Chen, Y. Y., & Liu, M., 2019. Tightly Coupled 3D Lidar Inertial Odometry and Mapping, in: 2019 International Conference on Robotics and Automation, Montreal. https://doi.org/10.1109/ICRA.2019.8793511

Zhang, J., & Singh, S., 2014. LOAM: Lidar odometry and mapping in real-time., in: Robotics: Science and Systems X, Berkeley. https://doi.org/10.15607/RSS.2014.X.007

Zhang, S., & Zhang, L. M., 2017. Impact of the 2008 Wenchuan earthquake in China on subsequent long-term debris flow activities in the epicentral area. Geomorphology, 276, 86-103. https://doi.org/10.1016/j.geomorph.2016.10.009

Zhang, S., Zhang, L. M., & Chen, H. X., 2014. Relationships among three repeated large-scale debris flows at Pubugou Ravine in the Wenchuan earthquake zone. Canadian Geotechnical Journal, 51(9), 951-965. https://doi.org/10.1139/cgj-2013-0368

Zhang, W., Chen, J. Q., Ma, J. H., Cao, C., Yin, H., Wang, J., & Han, B., 2023. Evolution of sediment after a decade of the Wenchuan earthquake: a case study in a protected debris flow catchment in Wenchuan County, China. Acta Geotechnica. https://doi.org/10.1007/s11440-022-01789-x

Zhang, X. Z., Tang, C. X., Li, N., Xiong, J., Chen, M., Li, M. W., & Tang, C., 2022. Investigation of the 2019 Wenchuan County debris flow disaster suggests nonuniform spatial and temporal post-seismic debris flow evolution patterns. Landslides, 19(8), 1935-1956. https://doi.org/10.1007/s10346-022-01896-6

Zhang, Y. Y., Huang, C., Huang, C., & Li, M. Y., 2022. Spatio-temporal evolution characteristics of typical debris flow sources after an earthquake. Landslides, 19(9), 2263-2275. https://doi.org/10.1007/s10346-022-01883-x

Zheng, C. R., Zhu, Q. Y., Xu, W., Liu, X. Y., Guo, Q. Z., & Zhang, F., 2022. FAST-LIVO: Fast and Tightly-coupled Sparse-Direct LiDAR-Inertial-Visual Odometry, in: 2022 IEEE/RSJ International Conference on Intelligent Robots and Systems, Kyoto. https://doi.org/10.1109/Iros47612.2022.9981107

Zhou, P., Tang, X. M., Wang, Z. M., Cao, N., & Wang, X., 2017. Vertical Accuracy Effect Verification for Satellite Imagery With Different GCPs. Ieee Geoscience and Remote Sensing Letters, 14(8), 1268-1272. https://doi.org/10.1109/Lgrs.2017.2705339